\documentclass{article}

\usepackage[preprint]{neurips_2021}

\usepackage[utf8]{inputenc} 
\usepackage[T1]{fontenc}    
\usepackage{xcolor}
\usepackage{amsmath}
\usepackage{subcaption}
\usepackage{graphicx}
\usepackage{algorithm} 
\usepackage[noend]{algpseudocode} 
\usepackage[colorlinks=true,urlcolor=blue,citecolor=green,linkcolor=red,bookmarks=true]{hyperref}
 \usepackage{comment}
\usepackage{wrapfig}

\usepackage{xcolor}         

\usepackage{url}            
\usepackage{booktabs}       
\usepackage{amsfonts}       
\usepackage{nicefrac}       
\usepackage{microtype}      

\usepackage{multirow}
\DeclareMathOperator*{\argmax}{argmax}
\title{InFlow: Robust outlier detection utilizing Normalizing Flows}

\author{%
  Nishant Kumar\thanks{Majority of the work done while at HZDR – Helmholtz-Zentrum Dresden-Rossendorf and TU Dresden} \\
  HZDR \\
  Dresden, Germany \\
  \texttt{n.kumar@hzdr.de} \\
  
  \And
  Pia Hanfeld\\
  CASUS \\
  Görlitz, Germany \\
  \texttt{p.hanfeld@hzdr.de} \\

   \And 
  Michael Hecht\\
  CASUS \\
  Görlitz, Germany \\
  \texttt{m.hecht@hzdr.de} \\ \\
  
    \And 
  Michael Bussmann\\
  CASUS \\
  Görlitz, Germany \\
  \texttt{m.bussmann@hzdr.de} \\
  
      \And 
  Stefan Gumhold\\
  CGV, TU Dresden \\
  Dresden, Germany \\
  \texttt{stefan.gumhold@tu-dresden.de} \\
  
        \And 
  Nico Hoffmann\\
  HZDR \\
  Dresden, Germany \\
  \texttt{n.hoffmann@hzdr.de} \\

}

\begin{document}

\maketitle

\begin{abstract}
Normalizing flows are prominent deep generative models that provide tractable probability distributions and efficient density estimation. However, they are well known to fail while detecting Out-of-Distribution (OOD) inputs as they directly encode the local features of the input representations in their latent space. In this paper, we solve this overconfidence issue of normalizing flows by demonstrating that flows, if extended by an attention mechanism, can reliably detect outliers including adversarial attacks. Our approach does not require outlier data for training and we showcase the efficiency of our method for OOD detection by reporting state-of-the-art performance in diverse experimental settings. \footnote{Code available at \url{https://github.com/ComputationalRadiationPhysics/InFlow}.}  
\end{abstract}

\section{Introduction}

Rapid advancement in imaging sensor technology and machine learning (ML) techniques has led to notable breakthroughs in several real-world applications. ML models typically perform effectively when the training and testing data are sampled from the same distribution. However, when applied to input data that are not similar to the training data, i.e. when they are far away from the training data distribution (e.g. OOD), these models can fail and the predictions of the model are not reliable anymore. This limitation prevents the safe deployment of these ML models in life-sensitive and real-world setups like autonomous driving and medical diagnosis. In these setups, plenty of OOD data naturally occurs due to various factors such as different image acquisition settings, noise in image scenes, and varied camera parameters. Therefore, a reliable deployment of an ML model requires that the model can detect anomalies so that these models do not provide high confidence predictions to such inputs.

 Deep generative models are commonly used for OOD detection in an unsupervised setting because of their ability to approximate the density of in-distribution samples as a probability distribution. It allows these models to assign
 lower likelihood to OOD inputs, rendering such inputs less likely to have been sampled from the in-distribution training set.  Generative models such as Normalizing flows \citep{ref1}; \citep{ref2}; \citep{ref3}; \citep{ref4}; \citep{ref5}; \citep{ref6} are especially suitable candidates for OOD detection as they provide tractable likelihoods. Let us define \({X} \in \mathcal{X} \cong \mathbb{R}^n\) as a random variable with input observations $x := (x_1,x_2,...,x_n)$ and  probability distribution \({x} \sim p({x})\) while \({Z} \in \mathcal{Z} \cong  \mathbb{R}^n\) as the random variable with latent observations $z := (z_1,z_2,...,z_n)$ and probability distribution \({z} \sim p({z})\). Now, according to the change of variables formula, we can define a series of invertible bijective mappings \(f: \mathcal{X} \longrightarrow  \mathcal{Z}\) where $f := (f_1,..,f_j,...,f_K)$ with parameters $\theta := (\theta_1,..,\theta_j,...,\theta_K)$ and \(j \in {1, \dots, K}\) being the number of coupling blocks to get ${z} = f(x) = f_K(f_{K-1}(...(f_j(...(f_2(f_1(x)))))))$. Therefore, the log-likelihood of the posterior distribution \(p_{\theta}({x})\) is given as,


\begin{figure}[!htb]
\begin{subfigure}{.333\textwidth}
  \includegraphics[width=\linewidth]{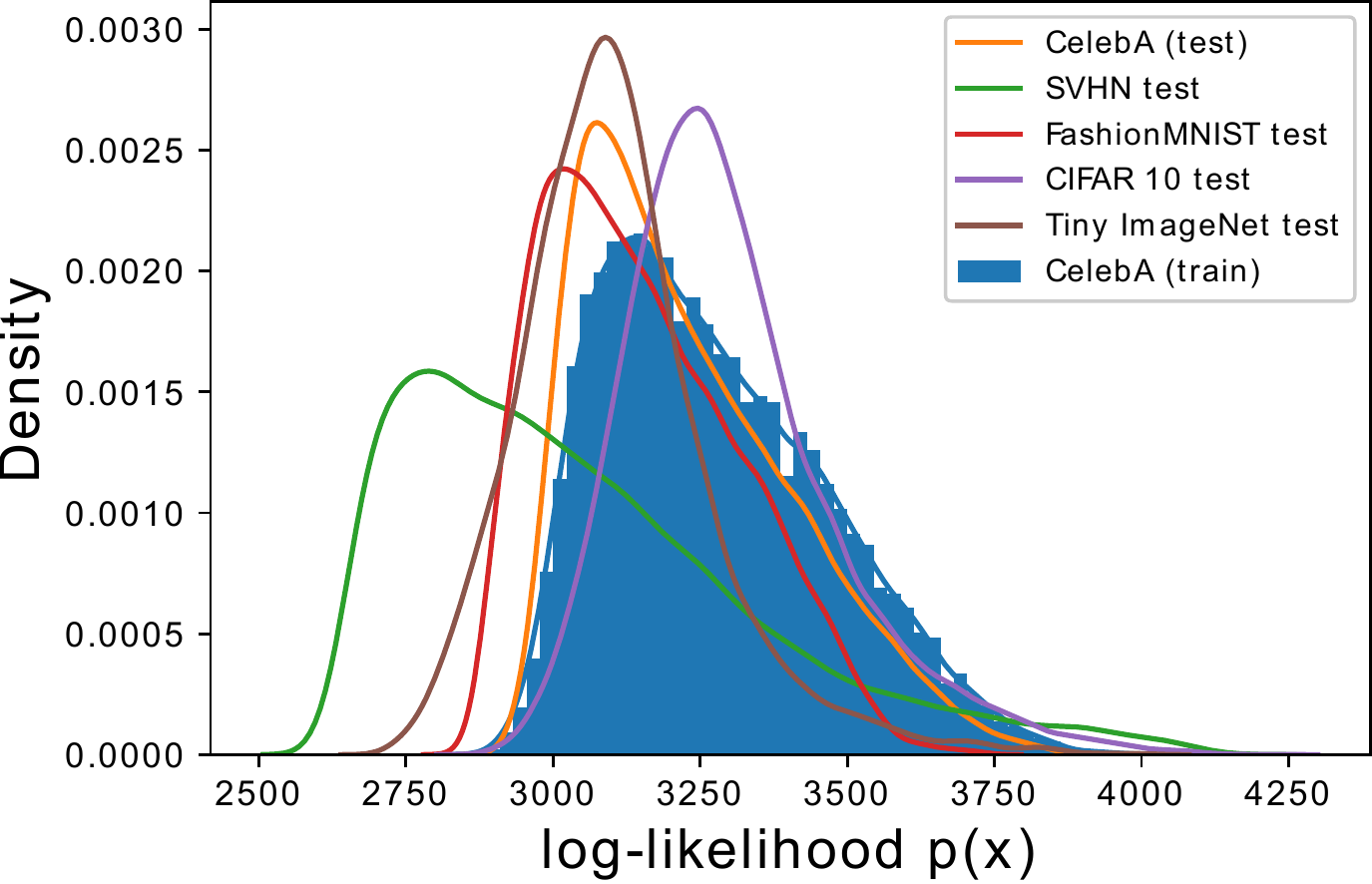}
  \caption{untrained RealNVP}
\end{subfigure}%
\begin{subfigure}{.333\textwidth}
  \includegraphics[width=\linewidth]{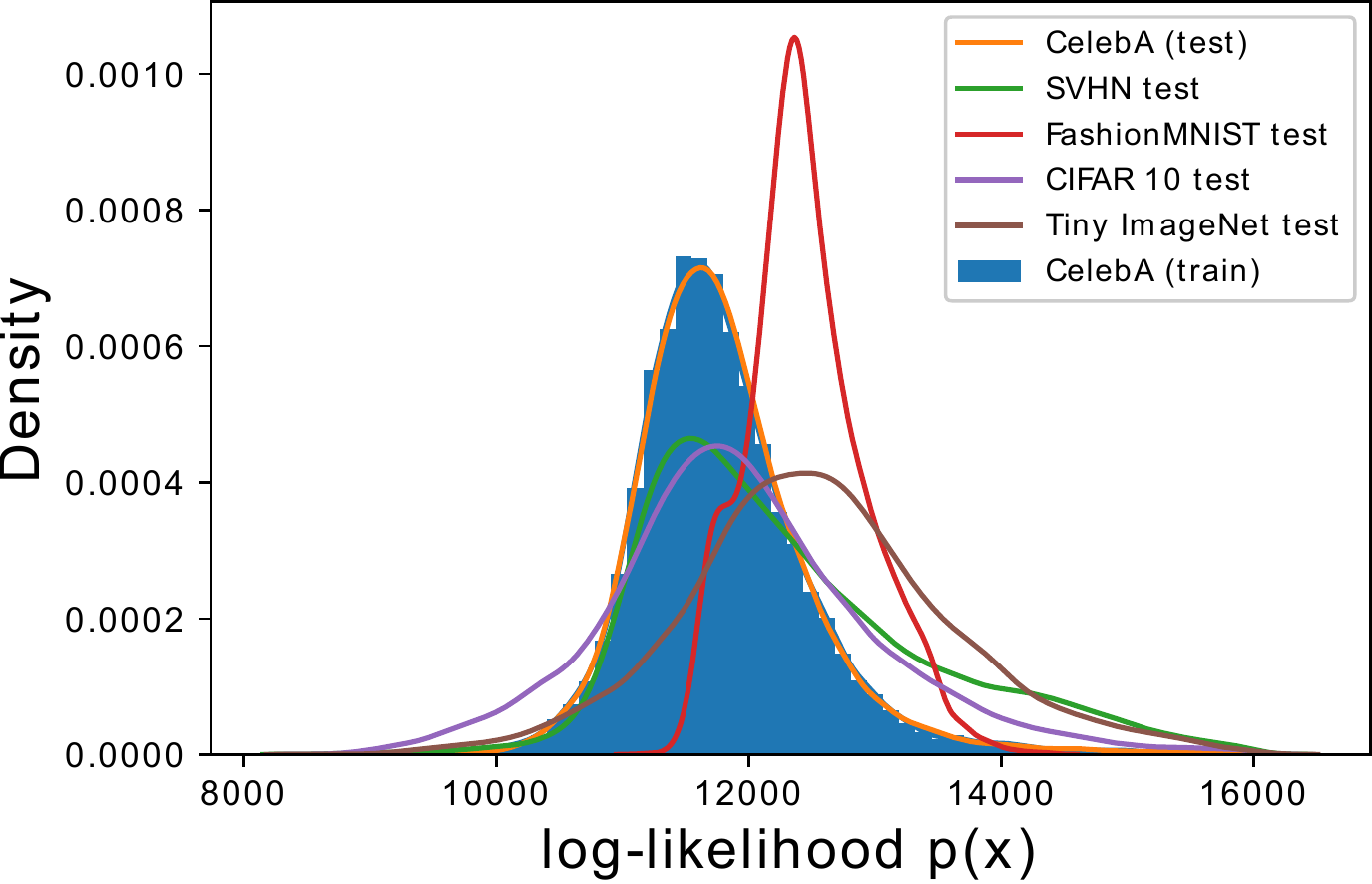}
  \caption{trained RealNVP}
\end{subfigure}%
\begin{subfigure}{.333\textwidth}
  \includegraphics[width=\linewidth]{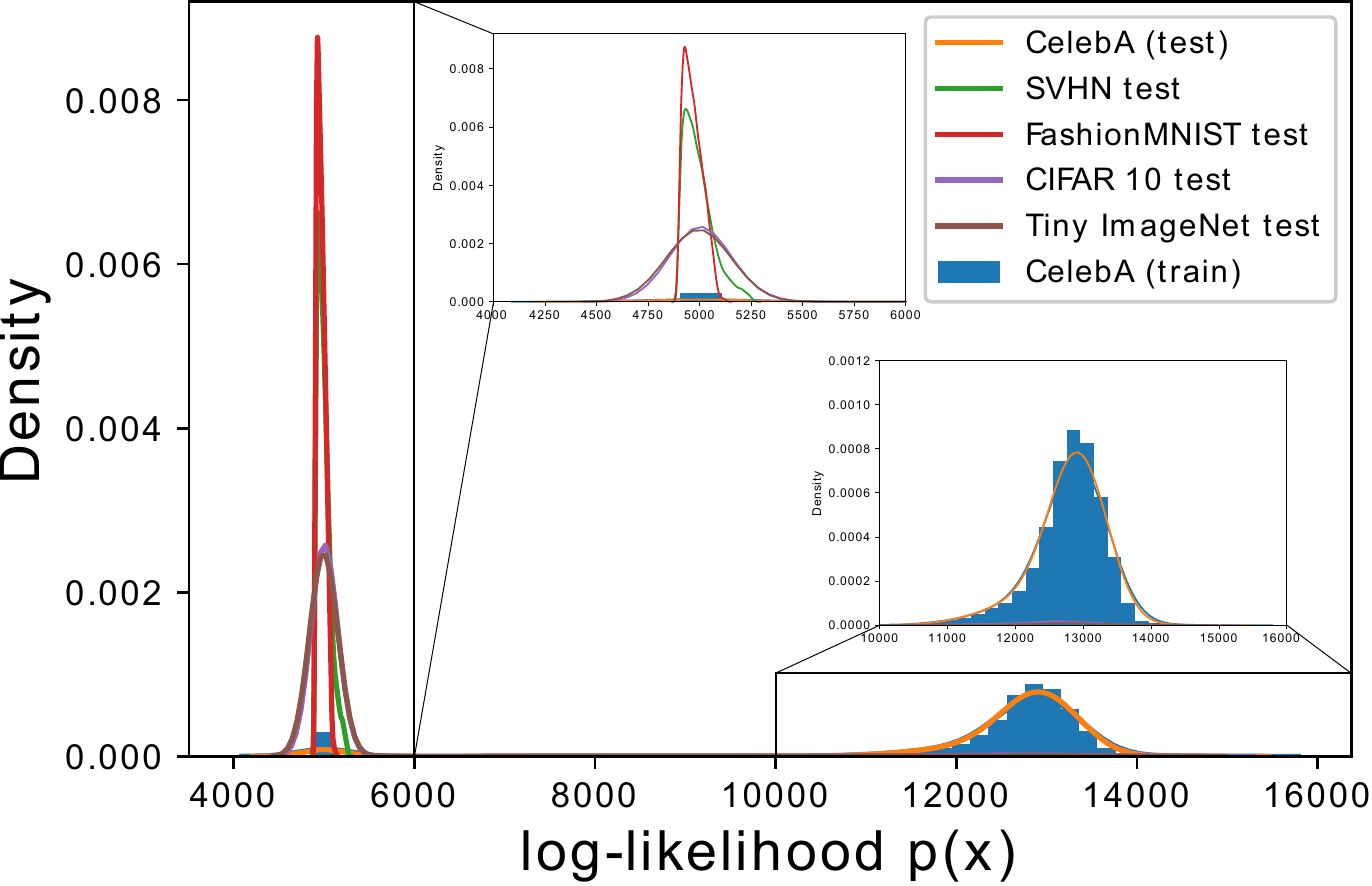}
  \caption{our InFlow model}
\end{subfigure}%
\caption{(a): A histogram of log-likelihoods of in-distribution CelebA and other OOD datasets for RealNVP model \citep{ref2} initialised with zeros. (b): A histogram of log-likelihoods for RealNVP model after training. Both (a) and (b) shows that RealNVP assigns higher likelihood to OOD inputs. (c): A histogram of log-likelihoods for our InFlow model at p-value $\alpha = 0.05$ assigning much higher log-likelihood to in-distribution CelebA samples than all other OOD datasets.}
\label{firstfigure}
\end{figure}

\begin{equation}
\log(p_{\theta}({x})) =  \log p(f(x)) + \log
  \left|
    \mathrm{det} \frac{
      \partial f(x)
    }{
      \partial {x}^T\
    }
  \right| \label{eq1} \end{equation}

Considering the prior distribution of latent space \(p({z})\) be a multivariate Gaussian, then the series of invertible bijective transformations with parameters $\theta$ can transform the posterior $\log(p_{\theta}({x}))$ from a Gaussian prior into significantly more complex probability distribution.  Hence, we can maximize the log-likelihood \(\log(p_{\theta}({x}))\) of the in-distribution samples \({x}\) with respect to the parameters $\theta$ of the invertible transformation 
$f$ and use a likelihood-based threshold to decide whether the log-likelihood of a test sample \({x}\) is below the threshold (classify $x$ as OOD) or above the threshold (classify $x$ as in-distribution). Additionally, AUCROC (Area Under the Curve Receiver Operating Characteristic) can be calculated to determine the performance of the flow model in terms of OOD detection. However, works such as \citep{ref7}; \citep{ref8} showed that generative models such as normalizing flows assign higher likelihoods to OOD samples compared to in-distribution samples, resulting in overconfident predictions on these OOD inputs as shown in Figure \ref{firstfigure} (b). To interpret this behavior, \citep{ref8} argued that these models only capture low-level statistics such as local pixel correlations rather than high-level semantics, due to which these models are inefficient in separating in-distribution data from the OOD samples. 

In this paper, we show that this issue can be solved by extending the normalizing flow design with an attention mechanism and validate that the attention mechanism ensures a higher log-likelihood score for in-distribution samples than the log-likelihood scores of OOD samples. 
We suggest that there are few benefits of constructing new designs of normalizing flow models for the OOD detection task and the focus should be directed towards extending the existing flow models with robust attention mechanisms in order to develop a reliable OOD detector. 
In Section \ref{relatedwork}, we present the current state of research in the field of OOD detection with the main focus on deep generative models. In Section \ref{body}, we develop the representation of our model and provide theoretical evidence for the robustness of our approach along with any underlying assumptions. In Section \ref{results}, we conduct several empirical evaluations of our approach in a variety of settings and discuss its effectiveness for OOD detection along with relevant limitations. 

\section{Related work}
\label{relatedwork}
 \citep{ref10} provided initial evidences that ML models have high confidence for OOD inputs. To overcome this issue, \citep{ref11} presented a likelihood ratio approach for OOD detection using auto-regressive generative models and experimented with a genomics dataset. \citep{ref12} employed input data perturbations to obtain a softmax score from a pre-trained model and used a threshold to determine whether the input data is in-distribution or OOD. \citep{ref13} modified a pre-existing network architecture and added a confidence estimate branch at the penultimate layer to enhance the OOD detection accuracy. \citep{ref14} applied a technique called outlier exposure that teaches a pre-trained model to detect unseen OOD examples. \citep{ref15}; \citep{ref45}; \citep{ref46} proposed classification models to detect OOD inputs whereas  \citep{ref29} utilized a combination of dimensionality reduction techniques and robust test statistics like Maximum Mean Discrepancy (MMD) to develop a dataset drift detection approach. \citep{ref16} proposed a confidence estimate based on Mahalanobis distances. \citep{ref17} showed that many existing OOD detection approaches such as \citep{ref12}; \citep{ref15}; \citep{ref16} do not work efficiently when small perturbations are added to the in-distribution samples. Hence, they trained their model on adversarial examples of in-distribution data along with the distribution from the outlier exposure developed by \citep{ref14}.
 \citep{ref47} defined their OOD detection strategy based on the idea that Generative adversarial networks (GANs) will not reconstruct OOD samples well.  \citep{ref20} developed a training mechanism by minimizing Kullback-Leibler (KL) divergence loss on the predictive distributions of the OOD samples to the uniform ones providing a measure for confidence assessment. \citep{ref21} used a self-supervised learning approach that is robust to detecting adversarial attacks while \citep{ref22} showed that the likelihood scores from generative models have a bias towards the complexity of the input data where non-smooth images tend to produce low likelihood scores while the smoother samples produce higher likelihood scores. \citep{ref25} studied OOD detection for Variational Auto-Encoders (VAEs) and proposed a likelihood regret score that computes the log-likelihood improvement of the VAE configuration that maximizes the likelihood of an individual sample. \citep{ref26} did not use likelihood-based OOD detection but utilized kernel density estimators such as Support Vector Machines (SVM) to differentiate between in-distribution and anomalous inputs. \citep{ref27} mined informative OOD data to improve the OOD detection performance, and subsequently generalized to unseen adversarial attacks.  \citep{ref30} showed that the high likelihood behavior of generative models for OOD samples is due to a mismatch between the model's typical set and its high probability density whereas \citep{ref33} introduced Watanabe–Akaike information criterion (WAIC) based score to differentiate OOD samples from in-distribution samples. \citep{ref23} gave an outline of several normalizing flow-based methods and discussed their suitability for different real-world applications. \citep{ref7} showed that INNs are especially attractive for OOD detection compared to other generative models such as VAEs and GANs since they provide an exact computation of the marginal likelihoods, thereby requiring no approximate inference techniques. Inspired from the work of \citep{ref19}, \citep{ref18} utilized Information Bottleneck (IB) as a loss function for the Invertible Neural Networks (INNs) with RealNVP  \citep{ref2} architecture to provide high-quality uncertainty estimation and OOD detection. \citep{ref28} introduced a residual flow architecture for OOD detection that learns the residual distribution from a Gaussian prior.

\section{InFlow for OOD detection}
\label{body}
Given unlabeled in-distribution samples $x:= (x_1,x_2,...,x_n)$, the task is to develop a robust normalizing flow model that maximizes the log-likelihood of in-distribution $\log p_{\theta}(x)$ whereas assigning lower log-likelihoods to OOD test samples. For achieving this, we explore the answer to the following questions: i). how can the maximum likelihood-based objective of our attention-based normalizing flow assign a higher log-likelihood to the in-distribution data than the log-likelihood of unseen OOD outliers? (see Section \ref{att_nfs}); ii). how do we define the attention mechanism that makes the normalizing flow model robust? (see Section \ref{mmd}). iii). how do we estimate an effective likelihood-based threshold for classifying the test samples as in-distribution or OOD? (see Section \ref{oodthreshold}).

\subsection{Model definition}\label{att_nfs}

 \citep{ref2} presented a normalizing flow architecture that are based on a sequence of high dimensional bijective functions $f_j$ stacked together as affine coupling blocks. Each of the affine coupling blocks contain the transformations, scaling $s$ and translation $t$ respectively. We extend this design by forwarding a function $c(x)$ (see also Appendix \ref{model}) to each of the $K$ coupling blocks as,
 
 \begin{equation}
 \begin{aligned}
 z = f(x) = f_K \big(f_{K-1}\big(  \cdots \big(f_j \cdots \big(f_2\big(c(x),f_1 \big(c(x),x\big)\big)\big) \big) \big) \big)  
 \end{aligned}
 \label{derive}
 \end{equation}

For simplicity, let us assume $K=2$, then Eq. \ref{derive} can be represented as $f(x) = f_2(c(x),f_1(c(x),x))$. Now, according to the chain rule, the derivative of $f(x)$ w.r.t. $x$ is given as, 

\small	
\begin{equation}
\begin{aligned}
\dfrac{df(x)}{dx^T} = \dfrac{\partial f_2(c(x),f_1(c(x),x))}{\partial c(x)} \dfrac{dc(x)}{dx^T} + \dfrac{\partial f_2(c(x),f_1(c(x),x))}{\partial f_1(c(x),x)}  \Big[\dfrac{\partial f_1(c(x),x)}{\partial c(x)}  \dfrac{dc(x)}{dx^T} + \dfrac{\partial f_1(c(x),x)}{\partial x^T}\Big] \label{chain1}
\end{aligned}
\end{equation}
\normalsize

By defining function $c(x)$ as the attention mechanism, that maps the input $x$ to the two integers $\{0,1\}$ where $c(x) = 1$ if $x$ is in-distribution and $c(x) = 0$ otherwise, produces the derivative of $c(x)$ w.r.t. $x$ as 0 except at the decision boundary of $c(x)$. Hence, the Eq. \ref{chain1} becomes,

\begin{equation}
\begin{aligned}
\dfrac{df(x)}{dx^T}= \dfrac{\partial f_2(c(x),f_1(c(x),x))}{\partial f_1(c(x),x)}  \dfrac{\partial f_1(c(x),x)}{\partial x^T} \label{chain2}
\end{aligned}
\end{equation}
 
It is observable that each of the derivatives in Eq. \ref{chain2} are the partial derivatives of the output of a single coupling block with respect to the input of the same coupling block. Hence, defining ${u_j}$ as the input and $v_j$ as the output of the $j^{th}$ coupling block and extending the Eq. \ref{chain2} with $K$ coupling blocks will lead to,

\begin{equation}
\begin{aligned}
z = \dfrac{df(x)}{dx^T}= \dfrac{\partial v_K}{\partial u_K^T}  \dots \dots  \dfrac{\partial  v_j}{\partial u_j^T} \dots \dots \dfrac{\partial  v_1}{\partial x^T} \label{chain3}
\end{aligned}
\end{equation}
 
At every affine coupling block, ${u_j}$ is channel wise divided into two halves ${u_{1j}}$ and ${u_{2j}}$ and $u_{1j}$ is transformed by the affine functions $s$ and $t$ respectively. We now multiply $c(x)$ with the output of transformations $s$ and $t$ in each of the coupling blocks. Therefore, the $j^{th}$ coupling block of our model is denoted as:  
\begin{equation} 
\begin{aligned}
    v_{1j} =  u_{1j}  \quad  ; \quad
        v_{2j} = u_{2j} \odot \exp(s(u_{1j}) * c(x)) + t(u_{1j}) * c(x)
\end{aligned} \label{eq3}
\end{equation}
where $v_{1j}$ is one part of the output $v_j$ which is replicated from input $u_{1j}$ and $v_{2j}$ is the other part which is the result of applying affine transformations on $s(u_{1j})$ and $t(u_{1j})$ respectively. Therefore, the jacobian matrix $\frac{
      \partial v_j
    }{
      \partial u_j^T\
    }$ at $j^{th}$ coupling block is given as:

\begin{equation} 
\frac{
      \partial v_j
    }{
      \partial u_j^T\
    }
=
\begin{bmatrix}
    \mathbb{I} & 0 \\
    \frac{
      \partial v_{2j}
    }{
      \partial u_{1j}^T\
    } & exp(s(u_{1j}) * c(x))
\end{bmatrix} \label{eq4pt1}
\end{equation}

 As there is no connection between $u_{2j}$ and $v_{1j}$ while $u_{1j}$ is equal to $v_{1j}$, the jacobian matrix in Eq. \ref{eq4pt1} is triangular which means its determinant is just the product of its main diagonal elements. These main diagonal elements of the jacobian matrices at each coupling block is multiplied to obtain the determinant of our end-to-end InFlow model as:

\begin{equation} 
\left| det 
\frac{
      \partial f(x)
    }{
      \partial x^T\
    } \right |
=
exp \Big[ \sum_{j=1}^{K} s(u_{1j}) * c(x) \Big] \label{eq5pt1}
\end{equation} 

Since the attention mechanism $c(x)$ is a common element, applying logarithm on Eq. \ref{eq5pt1} gives:

\begin{equation} 
log \left| det 
\frac{
      \partial f(x)
    }{
      \partial x^T\
    } \right |
=
log \Big( \big(exp \sum_{j=1}^{K} s(u_{1j})\big)^{c(x)} \Big) = c(x) * \sum_{j=1}^{K} s(u_{1j}) \label{eq6pt1}
\end{equation} 

It is to be noted that the output $v_{j} = [v_{1j}, v_{2j}]$ is still invertible and the mappings $s$ and $t$ can be arbitrarily non-invertible functions such as deep neural networks.  Hence, the parameters $\theta$ of the model can be optimized by minimizing the negative log-likelihood of the posterior $\log p_\theta(x)$ which is equivalent to maximizing the evidence lower bound $\mathcal{L}_K(x; c(x) = 1; \theta)$. Therefore, the maximum likelihood objective of the in-distribution samples can be achieved using Adam optimization with gradients of the form $\nabla_{\theta}(\mathcal{L}_K(x ; c({x}) = 1; \theta))$ as,

\begin{equation}
\argmax_{\theta} { \mathcal{L}_K(x; c(x) = 1; \theta}) = \mathbb{E}_{p(x;c(x)=1;\theta)} \big[\log(p(f(x))) + \sum_{j=1}^{K} s(u_{1j})\big]
\label{mlobjective}
\end{equation}

\paragraph{Proposition:}
\textit{
Considering input samples $x$ and the attention based normalizing flow model  \(f: \mathcal{X} \longrightarrow  \mathcal{Z}\) that satisfies $c(x) = 0$, then the model returns the prior distribution $p(z)$ of the latent observation $z$ for the posterior distribution $p_{\theta}(x)$} 

\paragraph{Proof:}
Theoretically, we have to prove that $p(z) = p_\theta(x)$ for all $x$ that satisfies $c(x) = 0$. Using the change of variables formula, the forward direction of an invertible normalizing flow
is,
\begin{equation}
p(z) = p_\theta\Big(x = f^{-1}(z)\Big) \left|\det \frac{
      \partial f^{-1}({z})
    }{
      \partial z^T\
    }\right|
\label{eq5}
\end{equation}
Now, using $c(x) = 0$ in the Eq. \ref{eq3} will yield $v_{1j} =  u_{1j}$ and $v_{2j} = u_{2j}$. This conveys that the output $v_j$ of the $j^{th}$ coupling block is equal to the input $u_j$ considering $u_j = [u_{1j},u_{2j}]$ and $v_j = [v_{1j},v_{2j}]$. Therefore, by substituting output of each coupling block with its input, we get $f(x) = x$. Additionally, for the reverse transformation, the change of variables formula gives,
\begin{equation}
p_\theta(x) = p\Big(z = f(x)\Big) \left|\det \frac{
      \partial f({x})
    }{
      \partial x^T\
    }\right|
\label{eq6}
\end{equation}

Using the result that $f(x) = x$ for $c(x) = 0$ in Eq. \ref{eq6}, we will obtain:
\begin{align}
p_\theta(x) & = p(z)\left|\det \frac{
      \partial x
    }{
      \partial x^T\
    }\right| = p(z) \left|\det[I]\right| 
 = p(z)
 \label{eq7}
\end{align}

Therefore, Eq. \ref{eq7} shows that the proposition holds and provides an elegant proof. With the condition $c(x_{out}) = 0$ satisfied, the posterior log-likelihood of the OOD samples $x_{out}$ is given as:

\begin{equation} \log p_\theta({x_{out}})\Big|_{c(x_{out}) = 0} = \log p({z_{out}})  \label{eq9}
 \end{equation}

Furthermore, for the in-distribution samples $x_{in}$ that satisfies $c(x_{in}) = 1$, putting Eq. \ref{eq6pt1} in Eq. \ref{eq1} results in the  posterior log-likelihood of the in-distribution samples $x_{in}$ as,

  \begin{equation}  \log p_\theta({x_{in}}) \Big|_{c(x_{in}) = 1} =  \log p({z_{in}}) +  \sum_{j=1}^{K} s(u_{1j})
   \label{eq7pt1}
  \end{equation}

 Under the assumption that the maximum likelihood objective as shown in Eq. \ref{mlobjective} asymptotically converged, we argue that the empirical upper bound of $\log p({z_{in}})$ is equal to or larger than the maximum likelihood estimate (MLE) of $\log p({z_{out}})$, where MLE of $\log p({z_{out}})$ is attained when $x_{out} = z_{out} = 0$ (see Eq. \ref{MLE}) and $x_{out}$ is not transformed by the maximum likelihood training of our InFlow model. Additionally, in our implementation, $\sum_{j=1}^K s(u_{1j}) \geq 0$, since sub-networks $s$ and $t$ are realized by a succession of several simple convolutional layers with ReLU activations (see Table \ref{stnetwork} in Appendix \ref{model}). Considering these postulations, it is noticeable from Figure \ref{firstfigure} (c) that the log-likelihood of in-distribution samples $\log p_\theta({x_{in}})$ is significantly higher than the  log-likelihood of OOD samples $\log p_\theta({x_{out}}) $, leading to robust disentanglement of posterior log-likelihoods of in-distribution samples from the OOD samples.

\subsection{The attention mechanism}\label{mmd}

We utilize Maximum mean discrepancy (MMD) \citep{ref9} as our attention mechanism since it is an efficient metric to perform the two sample kernel tests. Assuming we have two distributions $P$ and $Q$ over the sets $\mathcal{X}$ and $\mathcal{Y}$ respectively, $k(.)$ as the kernel in a reproducing kernel Hilbert space (RKHS) given by $\mathcal{H}$ that maps $\varphi : \mathcal{X}, \mathcal{Y}\to \mathcal{H}$,   \({X} \in \mathcal{X} \cong \mathbb{R}^n\) be the input random variable with in-distribution observations $x := (x_1,x_2,...,x_n)$ where $X \stackrel{\mathit{iid}}{\sim} P$,   \({Y} \in \mathcal{Y} \cong \mathbb{R}^m\) be another random variable with unknown observations $y := (y_1,y_2,...,y_m)$ where $Y \stackrel{\mathit{iid}}{\sim} Q$, then the MMD in $\mathcal{H}$ between two distributions $P$ and $Q$ is given by,
\begin{equation}
MMD^2(\mathcal{H_K}, P, Q) = \lVert \mathbb{E}_{X \sim P}[ \varphi(X) ] - \mathbb{E}_{Y \sim Q}[ \varphi(Y) ]  \rVert_\mathcal{H_K}
\label{MMD}
\end{equation}
However, calculating $MMD^2(\mathcal{H_K}, P, Q)$ has quadratic time complexity due to which, given a subset of in-distribution observations $\bar{x}$ where $\bar{x} \subset x$, we use an encoder function $\phi : \bar{x}, y \rightarrow \hat{x}, \hat{y}$ that maps the high dimensional input space $x$ and $y$ into a lower $d$ dimensional space $\hat{x}$ and $\hat{y}$ with the new observations $\hat{x}:={x_1^d,x_2^d,...,x_n^d}$ and $\hat{y}:={y_1^d,y_2^d,...,y_m^d}$. The details related to the encoder architecture and the hyperparameters can be found in Appendix \ref{attentionmechanism}. Now, given the kernel $k(.)$, an unbiased empirical approximation of $MMD^2(\mathcal{H_K}, P, Q)$ on a lower $d$ dimensional space is a sum of two U-statistics and a sample average which is given by  \citep{ref9},
{\small
\begin{align}
  {MMD}_U^2 (\hat{x}, \hat{y}) = \frac{1}{n(n-1)} \sum_{i=1}^n \sum_{j\neq i}^n k({x}_i^d, {x}_{j}^d)
  + \frac{1}{m(m-1)} \sum_{i=1}^m \sum_{j \neq i}^m k({y}_i^d, y_{j}^d) 
  - \frac{2}{nm} \sum_{i = 1}^n \sum_{j = 1}^m k(x_i^d, y_j^d)
\label{eq12}
\end{align}
}%

We used ${MMD}_U^2 (\hat{x}, \hat{y})$ as a test statistic with the null hypothesis $H_0: P = Q$ while the alternate hypothesis being $H_1: P \neq Q$. Let us assume $\alpha$ be the significance p-value that gives the maximum permissible probability of falsely rejecting the null hypothesis $H_0$. Then under the permutation based hypothesis test, the set of all encoded observations i.e. $\hat{x} \cup \hat{y}$ is used to generate $P$ randomly permuted partitions with $(\hat{x}_p , \hat{y}_p)$ at $p = 1,2,...., P$. After performing the permutations, we compute ${MMD}_U^2 (\hat{x}_p, \hat{y}_p)$ for each instances of $p$ and compare it with ${MMD}_U^2 (\hat{x}, \hat{y})$ as presented in Algorithm \ref{alg2} of Appendix \ref{pseudocode}. We then calculate the mean p-value $\hat{\alpha}$ as the proportion of permutations $P$ where ${MMD}_U^2 (\hat{x}_p, \hat{y}_p) > {MMD}_U^2 (\hat{x}, \hat{y})$ holds. Finally, we reject our null hypothesis $H_0$ if $\hat{\alpha} < \alpha$ and define $c(y) = 0$ for the test samples $y$.

\subsection{Likelihood-based threshold for OOD detection}
\label{oodthreshold}
The decision for deep generative models to classify input test samples as in-distribution or OOD naturally grounds on the likelihood-based threshold. To realize a robust likelihood-based OOD detector, we assert that the minimum posterior log-likelihood score of an in-distribution sample should preferably be higher than the maximum posterior log-likelihood score of the OOD samples. Hence, we define our likelihood-based threshold for OOD detection as the maximum posterior log-likelihood of OOD samples $\mathcal{L}_{th} = \argmax  \log p(z_{out})$. Moreover, to study the effect of p-value $\alpha$ on the performance of our approach, we relate the significance p-value $\alpha$ with the confidence bounds of the Gaussian prior distribution and infer several critical values of this likelihood-based threshold based on $\alpha$. Therefore, $1- \alpha$ can be seen as the proportion of the data within the $w$ standard deviation of the mean $\mu$, with $w$ computable by the inverse of the error function, $er\!f^{-1}$ using, 

\begin{equation}
w = \sqrt{2} * er\!f^{-1}(1 - \alpha) 
\label{errorfunction} 
\end{equation}

Now, let us assume $l$ be the dimension of the latent observation $z_{out}$, then given the  mean $\mu$ and variance $(w\sigma)^2$ of the prior Gaussian distribution, the log-likelihood $\log p({z_{out}}; \mu; (w\sigma)^2 )$ is,

 \begin{equation}
 \log p(z_{out}; \mu; (w\sigma)^2)  =  - \frac{l}{2}\log(2\pi) - l \log(w\sigma) - \frac{1}{2(w\sigma)^2} \sum_{i=1}^{l} (z_{out}^l - \mu)^2
 \label{normal}
\end{equation}

The maximum likelihood estimate (MLE) of $\log p(z_{out}; \mu; (w\sigma)^2)$ can then be computed as the asymptotically unbiased upper bound that needs to satisfy the following condition,

 \begin{equation}
 \frac{\partial \log p(z_{out}; \mu; (w\sigma)^2)} {\partial z_{out}} = 0 = - \frac{1}{2(w\sigma)^2} \sum_{i=1}^{l} (z_{out}^l - \mu)^2
 \label{MLE}
\end{equation}

Given $\mu = 0$ and $\sigma^2 = 1$, the values of $z_{out}^l$ should be $0$ to satisfy the condition in Eq. \ref{MLE}. Hence, substituting this proposition in Eq. \ref{normal} yields:

 \begin{equation}
\mathcal{L}_{th} = \argmax_{z_{out}}  \log p(z_{out}; \mu; (w\sigma)^2)) =  - \frac{l}{2}\log(2\pi) - l \log(w\sigma)
\label{lowerbound}
\end{equation}

Eq. \ref{lowerbound} shows that the MLE upper bound $\mathcal{L}_{th}$ can be interpreted as a robust likelihood-based threshold since it is data-independent and only constrained on mean $\mu$, standard deviation $\sigma$ of the prior distribution as well as the p-value $\alpha$. Hence, for a fixed $\mu$ and $\sigma$, the critical values of threshold $\mathcal{L}_{th}$ can then be controlled by changing the significance p-value $\alpha$. Our likelihood-based threshold, $\mathcal{L}_{th}$, therefore enables us to interpret the robustness of our approach for OOD detection w.r.t. confidence level of our attention mechanism $c(x)$ given a p-value $\alpha$.

\section{Experimental Results}
\label{results}
We evaluated the performance of our method for its robustness in a variety of experimental settings. For all our experiments, we fixed an in-distribution dataset for training our InFlow model and inferred with several OOD datasets. The details related to the datasets used in our experiments can be found in Appendix \ref{dataset}. The particulars related to the hyperparameters used during training and inference are given in Appendix \ref{trainingdetails}. We intended to assess our approach by evaluating its robustness with three different types of outlier data categories. The first category of test samples is generated by adding different types of visible perturbations to the in-distribution data samples (see Appendix \ref{robustnessnoise}). The second category is related to adversarial attacks on the in-distribution data samples with invisible perturbations (see Appendix \ref{robustnessadv}). The third category is associated with the dataset drifts where the semantic information and object classes of the test dataset is unseen by our InFlow model during training. Under the third category, we present some of the results in Section \ref{quantcomparesota} while further results related to this category are shown in Appendix \ref{robustnessdrift}. We also visualized the sub-network $s$ and $t$ activations as well as the input $x$ and latent observations $z$ for in-distribution and OOD samples and compared the behavior of our InFlow model with that of a RealNVP model (see Appendix
\ref{visualization}).

\begin{table}[!htb]
  \caption{AUCROC values of our InFlow model at $\alpha = 0.05$ with CIFAR 10 training data as in-distribution samples compared with other OOD detection methods.}
  \label{cifar10}
  \centering
  \begin{tabular}{lllllll}
    \toprule                   
   Datasets  & InFlow     & Likelihood Ratio     &  LR & ODIN & Outlier exposure & IC  \\
    \midrule
   MNIST               &  $\textbf{1}$      &  0.961   & 0.996     &  0.997 & 0.999 &  0.991           \\
   FashionMNIST        &  $\textbf{1}$      &  0.939   & 0.989     &  0.995 & 0.995 &   0.972          \\
    SVHN               & $\textbf{1}$   &  0.224   & 0.763    &  0.970 & 0.983 &     0.919        \\
    CelebA             &  $\textbf{1}$      & 0.668    & 0.786     &  0.965 & 0.858 &   0.677          \\
    CIFAR 10 (train)    & 0.513              & 0.497    & 0.494     &  0.702 & 0.504 &   0.497          \\ 
    CIFAR 10 (test)    & 0.529              & 0.500    & 0.496     &  0.706 & 0.500 &   0.500          \\
    Tiny ImageNet      &  0.556             &  0.273   & 0.848     &  $\textbf{0.941}$ & 0.984 &   0.362           \\
    Noise       &  $\textbf{1}$             &   0.618  &   0.739   &    1    & 0.995 &  0.878          \\
    Constant           &  $\textbf{1}$      &  0.918   &   0.935   &    0.908    &     0.999 &   1        \\
    \bottomrule
  \end{tabular}
\end{table}

\paragraph{Metrics:}
We used three different metrics namely Area under the Curve-Receiver Operating Characteristic (AUCROC$\uparrow$), False Positive Rate at 95\% True Positive Rate (FPR95$\downarrow$) and Area Under the Curve-Precision Recall (AUCPR$\uparrow$) to quantitatively evaluate the likelihood-based OOD detection performance of our method compared with other approaches. A receiver operating characteristic is a plot between the true positive rate (TPR) vs. the false positive rate (FPR) that shows the performance of the binary classification at different threshold configurations. We assign the binary label 1 as the ground truth for the log-likelihood scores obtained from the training in-distribution samples and the binary label 0 as the ground truth for the log-likelihood scores obtained from the test samples. AUCPR is the plot between the precision and recall with the same ground truth as AUCROC while FPR95 is the false positive rate when the true positive rate is at minimum 95\%. 

\begin{figure}[!htb]
\centering
\begin{subfigure}{.45\textwidth }
  \includegraphics[width=\linewidth]{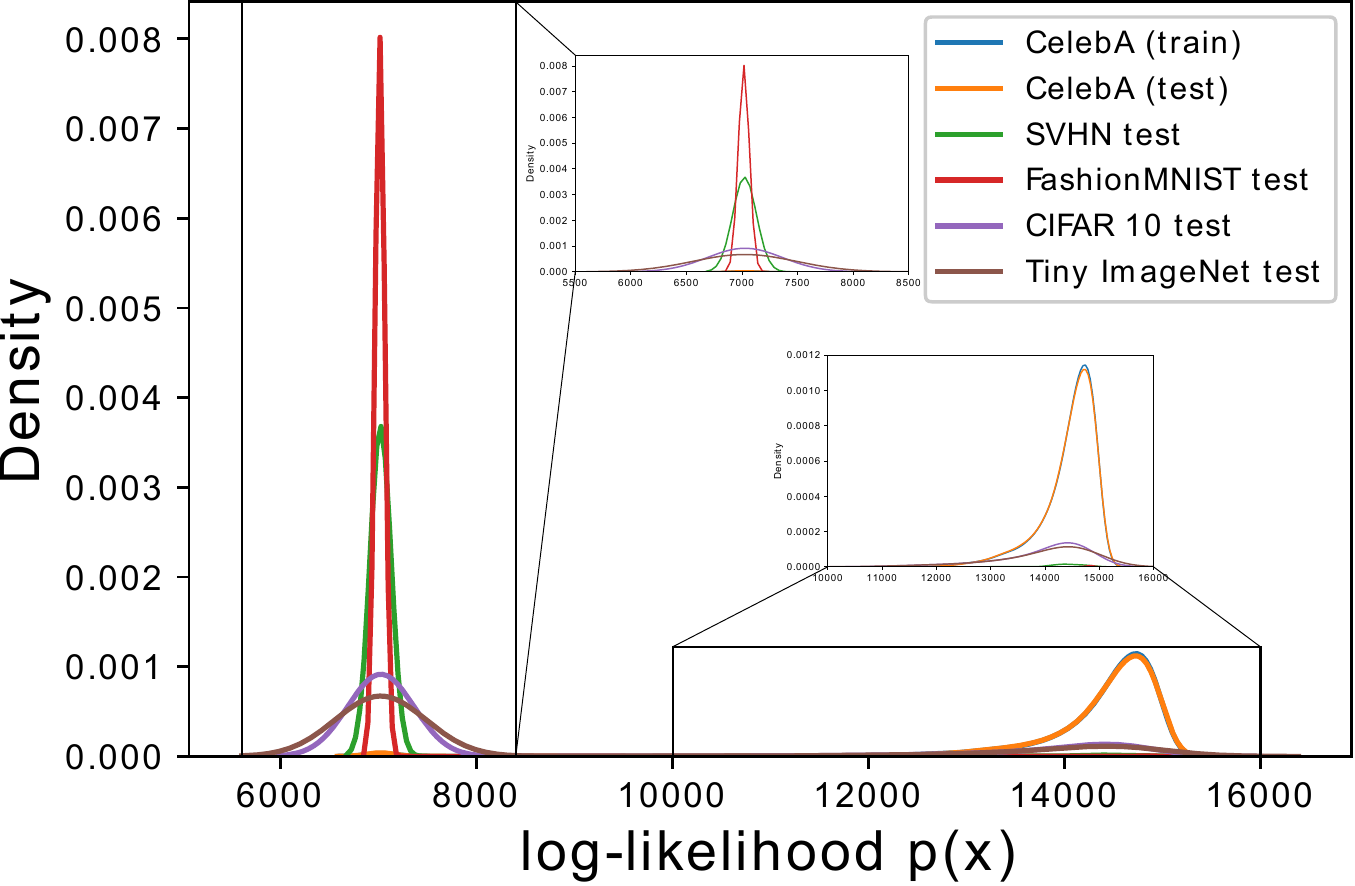}
  \caption{$\alpha = 10^{-4}$}
\end{subfigure}%
\begin{subfigure}{.45\textwidth}
  \includegraphics[width=\linewidth]{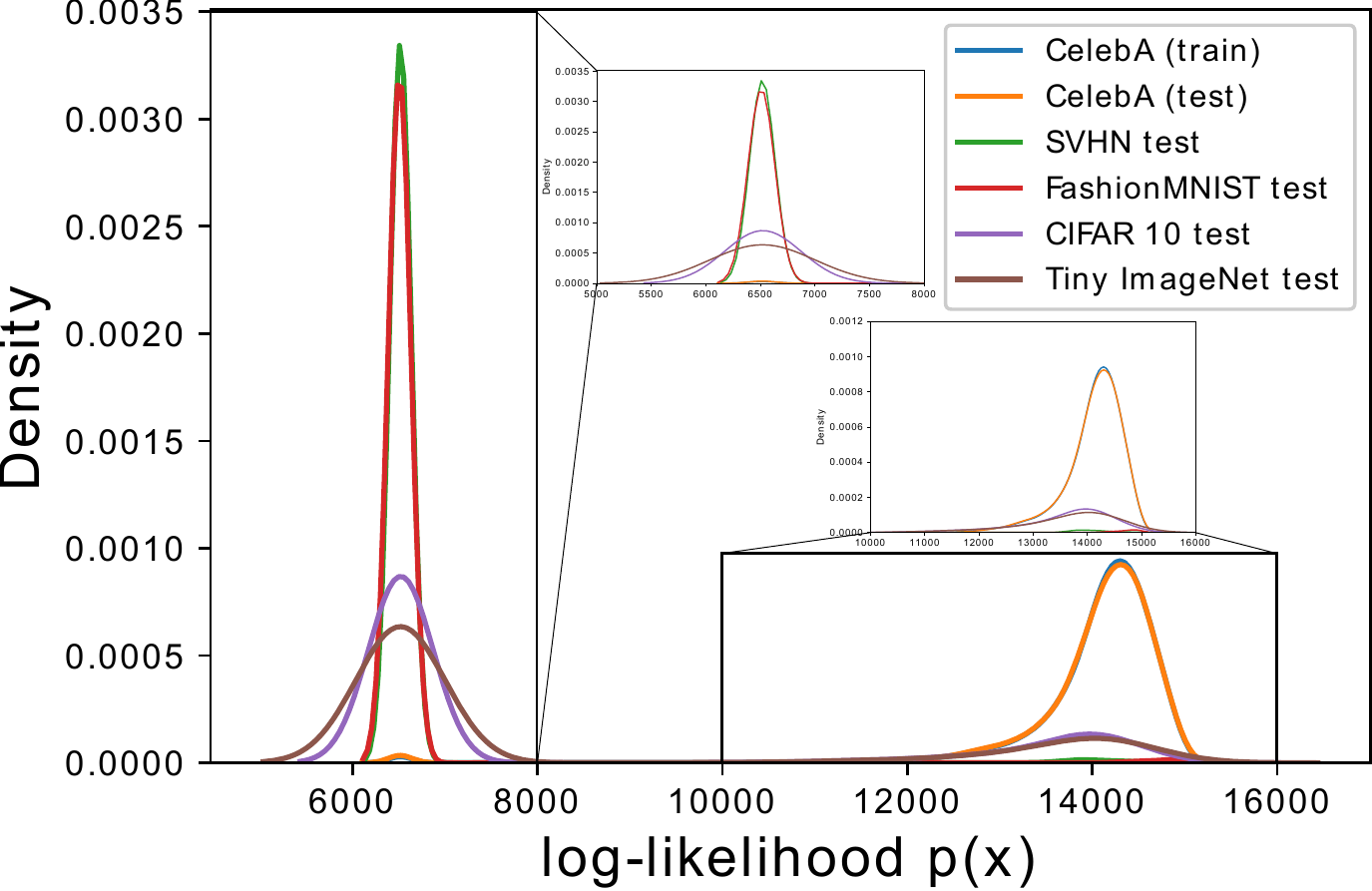}
  \caption{$\alpha =  10^{-3}$}
\end{subfigure}%
\\
\begin{subfigure}{.45\textwidth}
  \includegraphics[width=\linewidth]{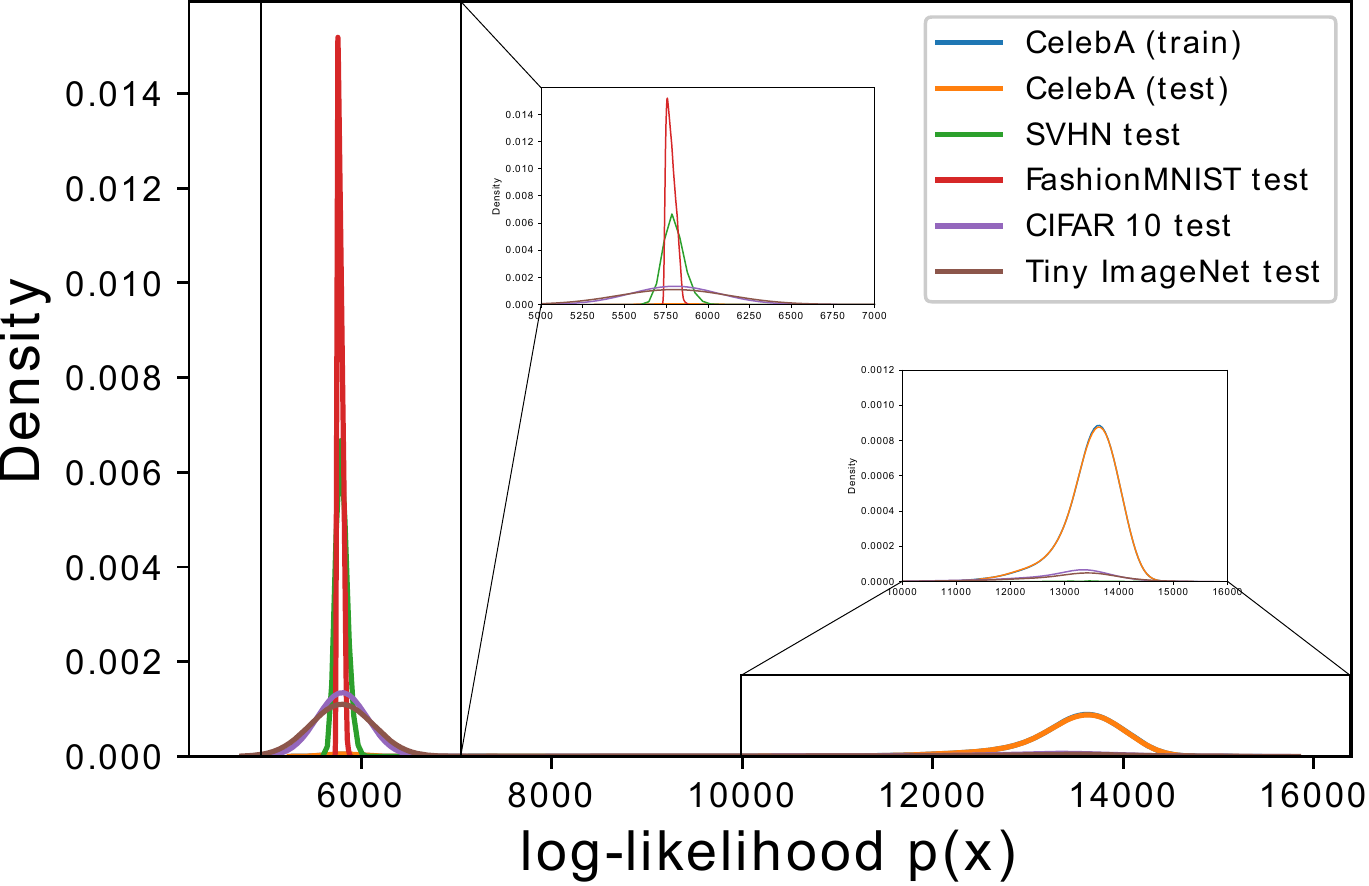}
  \caption{$\alpha =10^{-2}$}
\end{subfigure}%
\begin{subfigure}{.45\textwidth}
  \includegraphics[width=\linewidth]{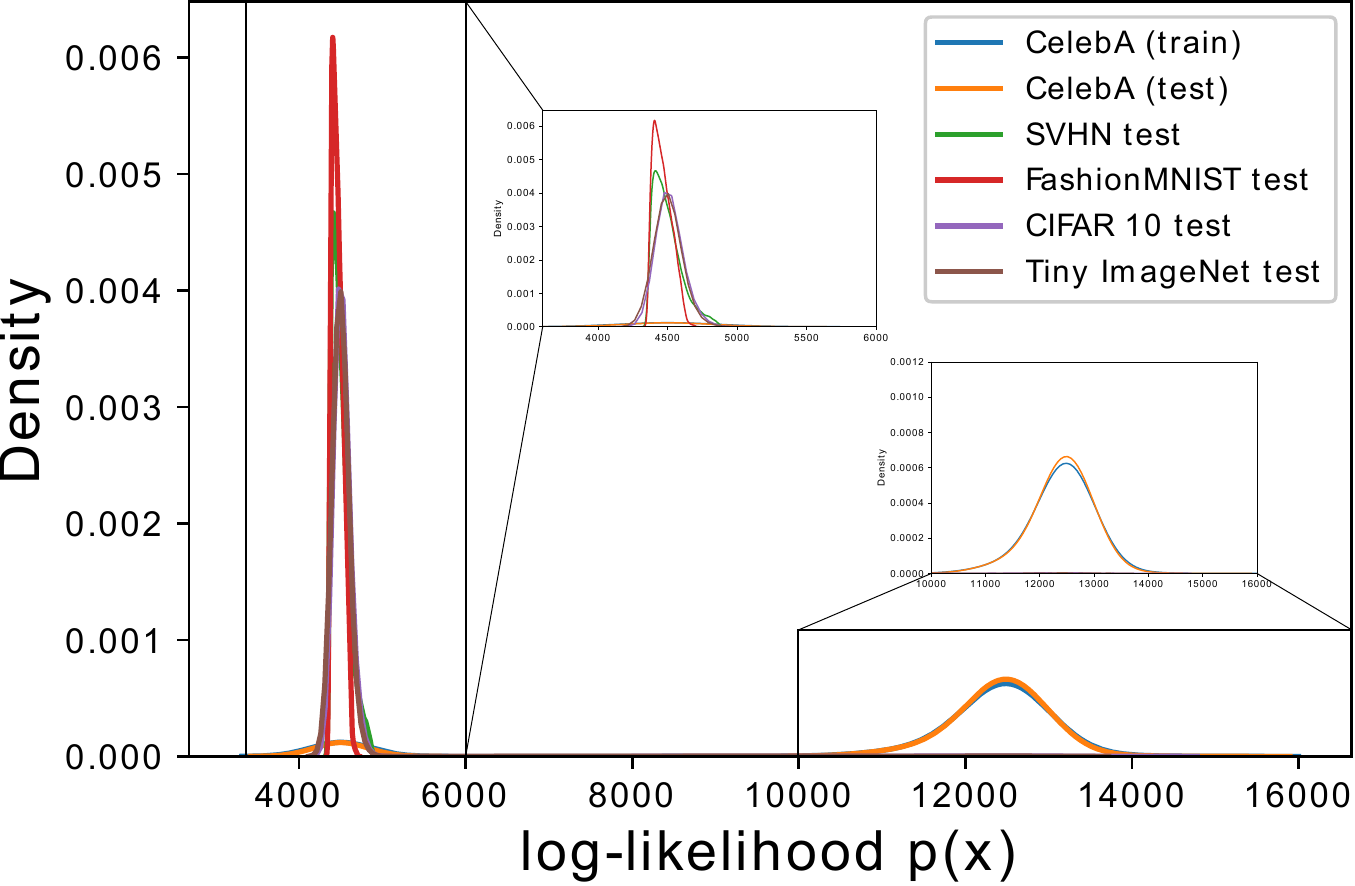}
  \caption{$\alpha = 10^{-1}$}
\end{subfigure}%
\caption{ The histogram of log-likelihoods of different datasets  at different values of significance p-value $\alpha$ when InFlow model was trained with CelebA training data.}
\label{alphat}
\end{figure}

\paragraph{Quantitative comparison with state-of-the-art:}
\label{quantcomparesota}
We study the performance of our InFlow model by comparing it with other likelihood-based OOD detection methods present in literature and compared the performance of our InFlow model with these methods using the three mentioned metrics. The methods that we evaluated are Likelihood ratio \citep{ref11} , Likelihood regret (LR) \citep{ref25}, ODIN \citep{ref12}, Outlier exposure \citep{ref14} and Input complexity (IC) \citep{ref22}. The details related to the implementation of these methods have been described in the Appendix \ref{sota}. Table \ref{cifar10} describes the AUCROC scores obtained from our model with CIFAR 10 training data as the in-distribution and compared with other approaches. It can be observed that except Tiny ImageNet test dataset, our model is robust and reaches the highest possible AUCROC scores in each of the evaluated OOD datasets. The  AUCROC scores of around 0.5 for CIFAR 10 training and test sets show that our model is unable to distinguish between the in-distribution CIFAR 10 samples which further verifies the robustness of our approach. Therefore, the results convey that our likelihood-based OOD detection is effective in solving the overconfidence issue of normalizing flows. The FPR95 and AUCPR scores for the same experimental setting are shown in Table \ref{cifar10sup} at Appendix \ref{robustnessdrift}.  The additional results for experiments related to dataset drift can be found in Table \ref{fmnist} at Appendix \ref{robustnessdrift} where we present the AUCROC, FPR95 and AUCPR scores for the evaluated methods with FashionMNIST as the in-distribution dataset. 

\paragraph{CIFAR 10 vs Tiny ImageNet:}
The results for the InFlow model trained with CIFAR 10 training data as shown in Table \ref{cifar10} reveals that our InFlow model has poor performance while detecting Tiny ImageNet test dataset as OOD. We associate this empirical outcome to two different rationale. Our first argument for such behavior relates to the significant overlap in the object class of these two datasets. It is to be noted that all 10 object classes of the CIFAR 10 testing samples are included in the object classes of Tiny ImageNet test set due to which the InFlow model assigns high log-likelihood scores to the test samples with overlapping classes in the Tiny ImageNet dataset. The second phenomenon is associated with the influence of image resolution on the log-likelihood score even if there is no class overlap between the samples from the two datasets. The test samples of CIFAR 10 are inherently $32 \times 32$ sized RGB images while Tiny ImageNet are of higher $64 \times 64$ resolution and are desirably downsampled to $32 \times 32$ to fit our experimental settings. We believe that decreasing the image resolution eliminated significant semantic information from the Tiny ImageNet samples that were important for OOD detection. Hence, we presume the resultant lower resolution Tiny ImageNet samples were of similar complexity compared to the CIFAR 10 samples. 

\begin{table}[!htb]
  \caption{AUCROC scores of our InFlow model trained on CelebA images as in-distribution and compared with different datasets at different significance p-value $\alpha$.}
  \label{alpha}
  \centering
  \begin{tabular}{llllll}
    \toprule  
   Datasets  &  $\alpha = 10^{-4}$ &  $\alpha =  10^{-3}$    & $\alpha = 10^{-2}$ & $\alpha = 5 \times 10^{-2}$ & $\alpha = 10^{-1}$ \\

    \midrule
  MNIST   &   1  &   1  &    1   &   1  & 1 \\
  FashionMNIST        &   0.986  &   0.986  &   1   &   1  & 1 \\
  SVHN                &   0.994  &   0.995  &   0.999   &  1 & 1 \\
  CelebA (train)       &   0.494 &   0.506   &  0.511  &   0.524   & 0.548 \\
  CelebA (test)       &   0.495 &   0.506   &  0.514   &   0.525   & 0.548 \\ 
   CIFAR10            &   0.931    &   0.930  &   0.965   &   0.990   & 0.998 \\
     Tiny ImageNet    &   0.926  &  0.925  &   0.969  &   0.993 & 0.998 \\
       Noise          &    1     &    1  &    1   &    1   & 1 \\
       Constant       &   0.999  &   0.999    &   1    &    1   &  1\\
  \bottomrule
  \end{tabular}
\end{table}

\paragraph{In/Out classification at the decision boundary:} We anticipate that the decision on whether a test sample is an in-distribution or OOD can change by adding extremely small and invisible perturbations to the test sample that lies at the decision boundary. These perturbations can be applied in the form of adversarial attacks and the OOD detection approach must be robust to such adversarial changes in the in-distribution samples. We performed exhaustive experiments for evaluating the robustness of our InFlow model w.r.t. such attacks. The results and the discussion related to it can be viewed in Appendix \ref{robustnessadv}. Our results convey that the usage of the MMD based hypothesis test as our attention mechanism is highly effective in detecting such adversarial changes since projecting the probability distribution of the attacked samples in higher dimensional RKHS space stretches its mean embeddings further away from the mean embeddings of the in-distribution samples.

\paragraph{p-value $\alpha$ and its limitations:}
\label{pvaluelimit}
We empirically evaluated the effect of p-value $\alpha$ on the robustness of our InFlow model for OOD detection. We trained the model on CelebA training data and inferred several datasets including CelebA with different significance p-values $\alpha$. Table \ref{alpha} shows AUCROC scores obtained for the evaluated datasets at different p-values ranging from $\alpha = 10^{-4}$ to $\alpha = 10^{-1}$. It can be observed that, in general, a smaller p-value $\alpha$ leads to a lower AUCROC score of the evaluated datasets. The visual evidence of this behavior is shown in Figure \ref{alphat} (a) where a number of OOD samples from CIFAR 10 and Tiny ImageNet test datasets attain high log-likelihood scores comparable to the log-likelihood scores of in-distribution samples. In contrast, a higher value of $\alpha$ leads to a number of in-distribution CelebA samples being wrongly classified as OOD. This is an apparent limitation of using p-value $\alpha$ as its dichotomy can significantly affect the decision-making of our InFlow model for OOD detection as no single p-value $\alpha$ can be interpreted as correct and foolproof for all types of data variability. This can lead to false positives and false negatives in life-sensitive real-world applications such as medical diagnosis and autonomous driving where the scope of failure is low. 

\section{Conclusion and Discussion}
\label{conclusion}

In this paper, we addressed the issue of overconfident predictions of normalizing flows for outlier inputs that have largely prevented these models to be deployed as a robust likelihood-based outlier detectors. With this regard, we put forth theoretical evidence along with exhaustive empirical investigation showing that the normalizing flows can be highly effective for detecting OOD data if the sub-network activations at each of its coupling blocks are complemented by an attention mechanism. We claim that considering the benefits of our approach, developing new flow architectures with high complexity particularly for OOD detection is not beneficial. In contrast, future work should instead focus on enhancing the attention mechanism to improve the robustness of these likelihood-based generative models for OOD detection. One approach in this direction is relating our OOD detection approach with a Generative adversarial network (GAN). To our understanding, the normalizing flow model can act as a generator network that learns to map the input samples into a latent space while the attention mechanism $c(x)$ can be viewed as a discriminator that distinguishes in-distribution samples from the OOD samples, thereby improving the performance of the generator for OOD detection.
The development of a robust OOD detection framework has a significant societal impact since such systems are crucial for the deployment of reliable and fair machine learning models in several real-world applications including medical diagnosis and autonomous driving. However, we urge caution while relying solely on OOD detection techniques for such sensitive applications and encourage more research in the direction of attention-based normalizing flows for OOD detection to further understand the limitations and mitigate potential risks. To the best of our knowledge, we are the first to overcome the high confidence issue of normalizing flows for OOD inputs and facilitated a methodological progress in this domain. 
\bibliographystyle{plainnat}


\newpage
\appendix

\section{Experimental Settings}
\label{appA}

\subsection{Datasets}
\label{dataset}

\begin{figure}[!htb]
 \centering
\begin{subfigure}{.24\textwidth}
  \includegraphics[width=\linewidth]{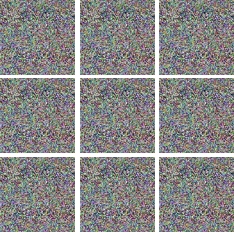}
  \caption{Noise}
\end{subfigure}%
  \hspace{0.2em}
\begin{subfigure}{.24\textwidth}
  \includegraphics[width=\linewidth]{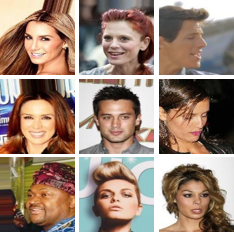}
  \caption{CelebA}
\end{subfigure}%
  \hspace{0.2em}
\begin{subfigure}{.24\textwidth}
  \includegraphics[width=\linewidth]{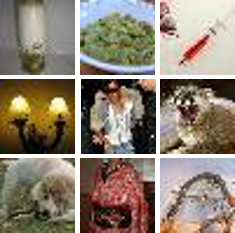}
  \caption{Tiny ImageNet}
\end{subfigure}%
  \hspace{0.2em}
\begin{subfigure}{.24\textwidth}
  \includegraphics[width=\linewidth]{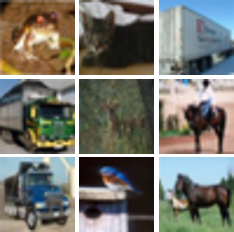}
  \caption{CIFAR 10}
\end{subfigure}%
\\
\begin{subfigure}{.24\textwidth}
  \includegraphics[width=\linewidth]{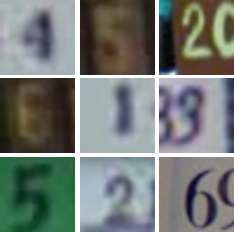}
  \caption{SVHN}
\end{subfigure}%
  \hspace{0.2em}
\begin{subfigure}{.24\textwidth}
  \includegraphics[width=\linewidth]{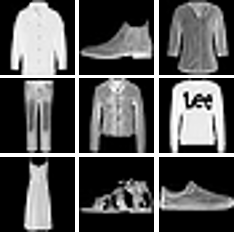}
  \caption{FashionMNIST}
\end{subfigure}%
  \hspace{0.2em}
\begin{subfigure}{.24\textwidth}
  \includegraphics[width=\linewidth]{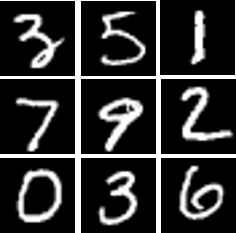}
  \caption{MNIST}
\end{subfigure}%
  \hspace{0.2em}
\begin{subfigure}{.24\textwidth}
  \includegraphics[width=\linewidth]{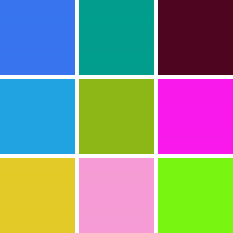}
  \caption{Constant}
\end{subfigure}%
  \caption{The resized RGB images from different datasets used in our experiments.}
  \label{fig2}
\end{figure}

We evaluated our model by carrying out experiments on publicly available datasets such as CelebA \citep{ref34}, MNIST \citep{ref35}, FashionMNIST \citep{ref36}, SVHN \citep{ref37}, CIFAR-10 \citep{ref38}, Tiny ImageNet \citep{ref39}. To maintain consistency, we created all input as  $32 \times 32 \times 3$ sized RGB images. Considering some of the evaluated datasets were of different resolution, we also resized those datasets as $32 \times 32 \times 3$ dimensional RGB images. For grayscale datasets such as MNIST and FashionMNIST, we concatenated the grayscale pixel values from the single-channel into three RGB channels. In addition to the publicly available datasets, we also synthetically generated two new datasets namely Noise and Constant to evaluate our method on the feature boundaries. For the Noise dataset, we performed a random sampling of integers between the range $[0,255]$ for each of the data points in all three RGB channels to obtain an RGB noise image while for the Constant dataset, we randomly sampled three different integers from the range $[0,255]$ and assigned it to each pixel of the three RGB channels respectively. Each of the evaluated datasets was normalized between the range $[0,1]$ before using it in our experiments. Table \ref{app:t} shows the original size of the datasets along with the number of images present in each of these datasets and the segregation of training and test sets. We keep the training set empty for all the datasets which were not used for training the InFlow model. 
Figure \ref{fig2} (a) - (h) shows nine examples of resized images in a $3 \times 3$ setting for each of the datasets with Noise being of highest feature complexity and Constant with the least.

\begin{table}[!htb]
  \caption{Details of the evaluated datasets such as size and the train-test partitions.}
  \label{app:t}
  \centering
  \begin{tabular}{llllll}
    \toprule  
     Dataset   & Actual size & Total images & Training set & Test set \\
    \midrule
  MNIST         &  $1 \times 28 \times 28$    &  70,000       &   60,000   &  10,000  \\
  FashionMNIST  &  $1 \times 28 \times 28$    &  70,000       &   -   &  10,000   \\  
  SVHN          &  $3 \times 32 \times 32$    &  99,289       &  -     & 26,032      \\
  CelebA &  $3 \times 178 \times 218$  &  202,599    &  150,000    &   52,599  \\ 
   CIFAR 10      &  $3 \times 32 \times 32$    &  60,000       &   50,000   &  10,000    \\
     Tiny ImageNet &  $3 \times 64 \times 64$   &  120,000    &  -    & 10,000   \\
       Noise &  $3 \times 32 \times 32$    &  10,000    &   -   &  10,000 \\
       Constant &  $3 \times 32 \times 32$    &  10,000    &   -   &  10,000 \\
    \bottomrule 
  \end{tabular}
\end{table}

\subsection{Model}
\label{model}
\begin{figure}[!htb]
 \centering
   \includegraphics[width=\linewidth]{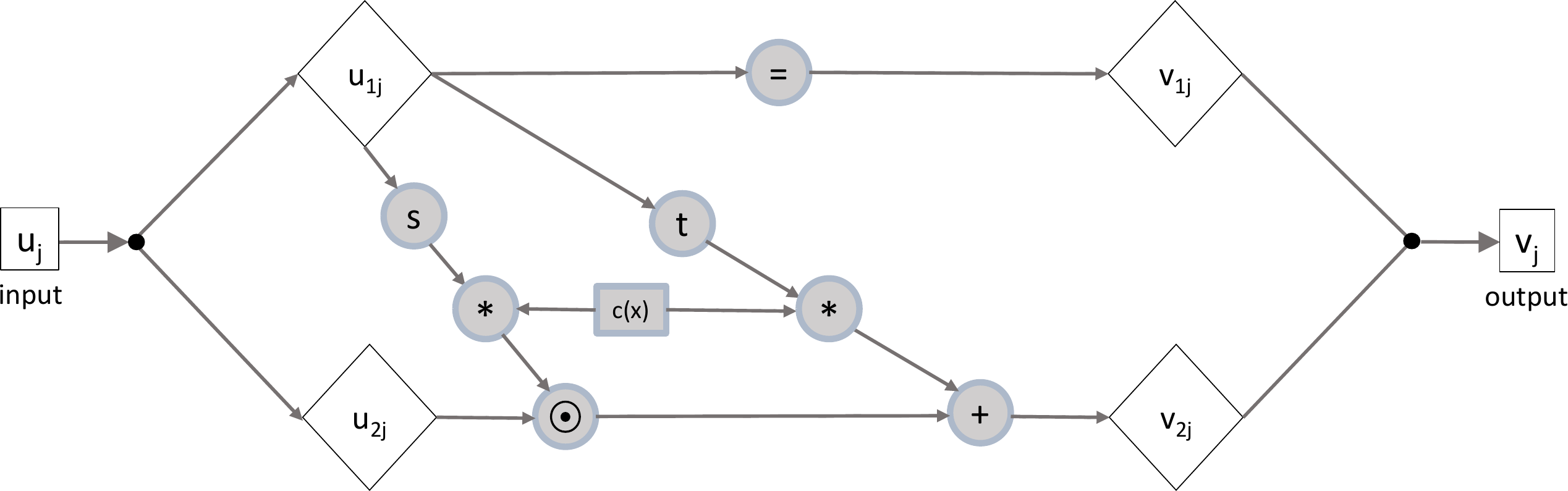}
     \caption{A single coupling block of our InFlow framework.}
  \label{couplingblock}
\end{figure}

As mentioned in Section \ref{att_nfs}, the central part of our normalizing flow model, InFlow, is an affine coupling block inspired by \citep{ref2}. Figure \ref{couplingblock} shows the architecture of our model at the $j^{th}$ coupling block with input $u_{j}$ and output $v_j$. The input $u_j$ was split channel-wise into two parts, $u_{1j}$ containing a single channel of the input RGB image and $u_{2j}$ with the remaining two channels.

The sub-part $u_{1j}$ is getting transformed with learnable functions $s$ and $t$ respectively, which we formulated as a neural network whose architectural details are in Table \ref{st}. The network consists of two convolutional layers with the ReLU unit as the non-linear activation function. The resolution of the input and output features in each of these convolutional layers is not changed.  As we defined ReLU as the last layer of our $s$ and $t$ sub-networks, we empirically ensured that the $\sum_{j}^{K} s(u_{1j}) \geq 0$.

\begin{table}[!htb]
  \caption{Network details for $s$ and $t$ transformations.}
  \label{stnetwork}
  \centering
  \begin{tabular}{llllll}
    \toprule  
   Operation  & In-channel   & Out-channel & Kernel & Stride & Padding \\
    \midrule
  Conv2D + ReLU  &  1   &  256   &  (3,3) & (1,1)   & (1,1)  \\  
  Conv2D + ReLU &  256   &  1    &  (3,3) & (1,1)    & (1,1)        \\ 
    \bottomrule 
    \label{st}
  \end{tabular}
\end{table}

\begin{figure}[!htb]
 \centering
   \includegraphics[width=\linewidth]{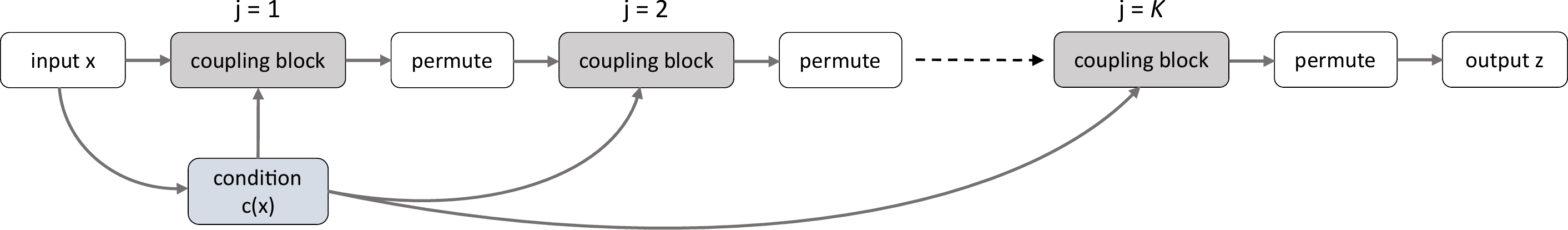}
  \caption{The end-to-end InFlow framework with $K$ coupling blocks.}
  \label{inflow}
\end{figure}

In addition to the use of learnable functions $s$ and $t$, we extended the design with an attention mechanism $c(x)$ by element-wise multiplying it with the output of the $s$ and $t$ networks as discussed in Section \ref{att_nfs}. Hence, each of these coupling blocks is stacked together to form our end-to-end InFlow framework as shown in Figure \ref{inflow}. We pass our attention mechanism $c(x)$ using a conditional node to each of the coupling blocks. Additionally, we perform random permutations of the variables between the two subsequent coupling blocks to ensure that the ordering of the sub-parts $u_{1j}$ and $u_{2j}$ are randomly changed across the channel dimension so that each channel is getting transformed using the $s$ and $t$ sub-networks at a particular coupling block of InFlow framework.

\subsection{Training details}
\label{trainingdetails}

We performed three different types of experiments for evaluating our model for robustness with OOD inputs as mentioned in Appendix \ref{OODevaluation}. We present the details related to the attention mechanism setup in Appendix \ref{attentionmechanism}. The InFlow model was trained in a comparable setting for each of the experiments where we used the Adam optimizer with initial learning rate of $1e^{-4}$, momentum $\beta_1 = 0.8$ and $\beta_2 =0.99$ and an exponential decay rate of $2e^{-5}$. Our model was trained on a single NVIDIA Tesla V100 GPU for 200 epochs with each epoch containing 100 training steps and a batch size of $250$ samples.

\subsection{The Pseudo Code}
\label{pseudocode}
The pseudo-code for training the InFlow model is described in Algorithm \ref{alg1}.
\begin{algorithm}
\caption{The maximum likelihood objective: InFlow}\label{compute_lle}
\begin{algorithmic}[1]
\Require In-distribution samples $x = (x_1, x_2,...,x_n)$, normalizing flow model $f$, number of iterations $i$, learning rate $\gamma$
\State  Choose a subset $\bar{x}$ with $\bar{x} \subset x$ and assign $c(\bar{x}) = 1$ 

\For{$i$ iterations}
\State $\theta \leftarrow \text{Adam}(\theta,\nabla_{\theta}(-\mathcal{L}_K(x ; c(\bar{x}) =1; \theta)),\gamma)$  \Comment{minimize negative log-likelihood}
\EndFor
\end{algorithmic}
\label{alg1}
\end{algorithm}

The pseudo-code for using  the  trained InFlow model for OOD detection is described in Algorithm \ref{alg2}.

\begin{algorithm}
\caption{OOD detection using InFlow model}\label{compute_lle}
\begin{algorithmic}[1]
\Require Unknown set of samples $y = (y_1, y_2,...,y_m)$, trained model $f$ and  p-value threshold $\alpha$. 
\State Use an encoder function $\phi: \bar{x},y \rightarrow \hat{x}, \hat{y}$  \Comment{Dimensionality reduction}
\State Take all encoded observations as ($\hat{x} \cup \hat{y}$) and perform $P$ permutations with $p = 1,..., P$.
\State Partition the set ($\hat{x} \cup \hat{y}$) into ($\hat{x}_p, \hat{y}_p$) for each of the $p$ permutations
\For{$p$ permutations} \Comment{Perform p-value permutation tests}

  \If{${MMD}_U^2 (\hat{x}_p, \hat{y}_p) > {MMD}_U^2 (\hat{x}, \hat{y})$} 
 \State assign the empirical p-value $\hat{\alpha}_p$ at $p^{th}$ permutation as 0 
 \Else \State assign the empirical p-value $\hat{\alpha}_p$ at $p^{th}$ permutation as 1
  \EndIf  
  \EndFor
\State Compute the mean p-value $\hat{\alpha}$ from the p-values $\hat{\alpha}_p$ for each of the $P$ permutations 
\If{$\hat{\alpha} < \alpha$}
\State reject null hypothesis $H_0$ and assign  $c($y$) = 0$
\State $\mathcal{L} = \mathcal{L}_K(y;c(y) = 0;\theta)$ \Comment{Estimate the log-likelihood}
\Else
 \State reject alternate hypothesis $H_1$ and assign $c(y) = 1$
\State $\mathcal{L} = \mathcal{L}_K(y;c(y) = 1;\theta)$ \Comment{Estimate the log-likelihood}
 \EndIf
 \State Estimate the likelihood-based threshold $\mathcal{L}_{th}$ with 
$w = \sqrt{2} * er\!f^{-1}(1 - \alpha)$ 
 \If{$\mathcal{L} < \mathcal{L}_{th}$}   
\State Assign $y$ samples as Out-of-Distribution (OOD) 
\Else
\State Assign $y$ samples as in-distribution 
 \EndIf
\end{algorithmic}
\label{alg2}
\end{algorithm}

\subsection{Implementing the state of the art}
\label{sota}

We quantitatively compared the robustness of our InFlow model with other popular OOD detection methods such as ODIN \citep{ref12}, Likelihood ratio \citep{ref11}, Outlier exposure \citep{ref14}, Likelihood regret (LR) \citep{ref25}, and Input Complexity (IC) \citep{ref22} using  AUCROC, FPR95 and AUCPR scores. During the evaluation, we fixed the same in-distribution samples for our approach as well as other competing methods. For ODIN, Outlier Exposure, and LR, we followed similar meta-parameter settings as recommended in the code documentation of these methods.  For the IC method, we followed the implementation as provided by \citep{ref25}. We first calculated the input data complexity using the length of the binary string provided by a PNG-based lossless compression algorithm and subtracted this input data complexity from the negative log-likelihood scores. For the Likelihood ratio method, we computed the scores by subtracting the log-likelihood of the background model from the log-likelihood of the main model. The background model was trained on perturbed input data by corrupting the input semantics with random pixel values. The number of pixels that were perturbed was $30\%$ of the total number of pixels for the model trained with FashionMNIST data and $20\%$ for the model trained with CIFAR 10.

\subsection{Evaluating robustness to OOD inputs}
\label{OODevaluation}
It is difficult to obtain a consensus in current literature upon a single definition for estimating the robustness of a machine learning model. Ideally, a model can be said to be robust if it is able to distinguish OOD inputs from the in-distribution samples. Therefore, it is essential to determine the true definition of an OOD outlier that a robust model should be able to detect. With this regard, we identify three such synopses which are generally used for assessing the robustness of an OOD detection approach,

\begin{itemize}

  \item \textbf{Dataset drift}: A robust OOD detection model should detect those input test samples that do not contain any of the object classes present in the in-distribution samples. These test samples have a complete shift in the semantic information of the data, and ideally, the model should not provide predictions with high confidence on such data since it has not observed such data during training (see Appendix \ref{robustnessdrift}).
  
  \item \textbf{Adversarial attacks}: A robust OOD detection model should be aware of adversarial attacks on the in-distribution samples. In these attacks, the magnitude of the perturbation is kept low so that the changes in the attacked sample are indistinguishable from the in-distribution samples but is enough to trick a model\textemdash e.g. a classifier\textemdash that interprets the attacked input sample with high confidence. Therefore, such attacks have the potential to significantly degrade the performance of ML models (see Appendix \ref{robustnessadv}). 
  
    \item \textbf{Visible perturbations}: We conduct another kind of robustness test where we corrupt the in-distribution samples with different types of perturbations.  Additionally, we test the performance of our model on different levels of corruption severity. The corruptions are visible for all severity levels even though the inherent semantic information is still preserved (see Appendix \ref{robustnessnoise}).
    
\end{itemize}

 \begin{table}
  \caption{Encoder network architecture for experiments without adversarial attacks.}
  \centering
  \begin{tabular}{llllll}
    \toprule  
   Operation  & In-channel   & Out-channel & Kernel & Stride & Padding \\
    \midrule
  Conv2D + ReLU &  3     &  64    &  (4,4) & (2,2)    & (0,0)  
  \\  
  Conv2D + ReLU  &  64    &  128      &  (4,4) & (2,2)    & (0,0)        \\ 
  Conv2D + ReLU  &  128    &  256      &  (4,4) & (2,2)    & (0,0)        \\ 
  Conv2D + ReLU  &  256    &  512      &  (4,4) & (2,2)    & (0,0)        \\ 
    \bottomrule 
 Operation &   In-features      &  Out-features         &        &          & \\
 \midrule
 Flatten +  Linear   &   2048    &    32       &   -     &    -     & -\\
    \bottomrule \\
\multicolumn{6}{c}{Encoder network architecture for experiments with adversarial attacks.}\\
    \toprule  
   Operation  & In-channel   & Out-channel & Kernel & Stride & Padding \\
    \midrule
  Conv2D + ReLU &  3     &  64    &  (4,4) & (2,2)    & (0,0)  
  \\  
  Conv2D + ReLU  &  64    &  128      &  (4,4) & (2,3)    & (0,0)        \\ 
  Conv2D + ReLU  &  128    &  256      &  (5,5) & (2,2)    & (0,0)        \\ 
  Conv2D + ReLU  &  256    &  512      &  (4,5) & (2,2)    & (0,0)        \\
  Conv2D + ReLU  &  512    &  512      &  (5,2) & (2,2)    & (0,0)        \\ 
    \bottomrule 
 Operation &   In-features      &  Out-features         &        &          & \\
 \midrule
 Flatten +  Linear   &   4096    &    32       &   -     &    -     & -\\
    \bottomrule    
    
    \label{encoder}
  \end{tabular}
\end{table}

\subsection{Dimensionality reduction}
\label{attentionmechanism}

MMD as a test statistic has considerable time and memory complexity for high-dimensional data. To overcome this challenge, we used an encoder to reduce the number of features per sample. The first part of Table \ref{encoder} shows the encoder network architecture used for the experiments that do not involve adversarial attacks. The second part of Table \ref{encoder} shows the encoder architecture used for experiments conducted for evaluating robustness to adversarial attacks. It can be observed that the final dimension obtained for samples consists of just 32 features for each of the encoder architectures. For MMD computation, we defined exponential quadratic function or Radial Basis function (RBF) as the kernel given by 
$k(x,y) = e^{-||x-y||^2/\sigma^2}$. 
The RBF kernel is positive definite due to which applying it on the input samples $x$ and $y$ that are dependent on variance $\sigma^2$ produce a smooth estimate in the RKHS space. This aids in better interpretation of the mean embeddings for the respective input distributions. For all the experiments involving permutation tests, the significance p-value $\alpha$  was set at $0.05$ and the number of permutations as $P= 100$. The average p-value $\hat{\alpha}$ was estimated for a batch of $250$ in-distribution samples and $50$ test samples.

\subsection{Robustness to dataset drift}
\label{robustnessdrift}

\paragraph{CelebA:}We utilized the CelebA training dataset as in-distribution and interpreted the influence of the p-value $\alpha$ on the OOD detection performance of our InFlow model. The AUCPR and FPR95  scores for such a setting are shown in Table \ref{alphafpraucpr}. It can be observed that as we lower the significance p-value $\alpha$, we obtain worse AUCPR and FPR95 scores, a behavior which is also discussed in Table \ref{alpha}. Since a smaller p-value $\alpha$ is statistically significant, a lower mean p-value $\hat{\alpha}$ is required to reject the null hypothesis that the test samples are from in-distribution. 
 
 \begin{table}[!htb]
  \caption{FPR95 and AUCPR values of our InFlow model trained on CelebA images as in-distribution and compared with different OOD datasets at different significance threshold $\alpha$ values.}
  \label{alphafpraucpr}
  \centering
  \begin{tabular}{llllll}
  \multicolumn{6}{c}{FPR95}\\
    \toprule
 Datasets  &  $\alpha = 10^{-4}$ &  $\alpha =  10^{-3}$    & $\alpha = 10^{-2}$ & $\alpha = 5 \times 10^{-2}$ & $\alpha = 10^{-1}$\\
    \midrule
  MNIST                &   0              &  0              &  0               &   0         & 0 \\
  FashionMNIST         &   0.010          &  0.010          &  0               &   0         & 0 \\  
  SVHN                 &   0.001          &  0.001          &  0               &   0         & 0 \\
  CelebA (train)       &   0.051          &  0.049          &  0.049           &   0.046      &  0.045 \\
  CelebA (test)        &   0.051          &  0.049          &  0.047           &   0.042      &  0.041 \\
  CIFAR 10             &   0.006         &  0.007           &  0.005           &   0.002     & 0 \\
  Tiny ImageNet        &   0.008           &  0.009         &  0.005           &   0.002     & 0 \\
  Noise                &   0               &  0             &  0               &   0         & 0 \\
  Constant             &   0               &  0             &  0               &   0         & 0 \\
  \multicolumn{6}{c}{AUCPR}\\
    \toprule                   
  Datasets  &  $\alpha = 10^{-4}$ &  $\alpha =  10^{-3}$    & $\alpha = 10^{-2}$ & $\alpha = 5 \times 10^{-2}$ & $\alpha = 10^{-1}$ \\
    \midrule
  MNIST               &  1               &   1                 &  1           &    1         &  1  \\
  FashionMNIST        &  0.990           &   0.990             &  1           &    1         &  1  \\  
  SVHN                &  0.996           &   0.996             &  0.999       &    1         &  1    \\
   CelebA (train)     &  0.488           &   0.517             &  0.531       &    0.559     & 0.599 \\ 
  CelebA (test)       &  0.487           &   0.517             &  0.530       &    0.556     & 0.599 \\ 
   CIFAR 10           &  0.970           &   0.970             &  0.986       &    0.996     & 0.999 \\
     Tiny ImageNet    &  0.912           &   0.910             &  0.966       &    0.993     & 0.998 \\
       Noise          &  1               &   1                 &  1           &    1         & 1\\
       Constant       &  0.999           &   0.999             &  1           &    1         & 1 \\
    \bottomrule 
  \end{tabular}
\end{table}

\paragraph{CIFAR 10:}We showed the efficiency of our InFlow model for detecting dataset drifts in Section \ref{results}. We further present the FPR95 and AUCPR scores (see Table \ref{cifar10sup}) for the setting where CIFAR 10 was fixed as in-distribution samples and the model was evaluated on other datasets. The interpretation of the achieved FPR95 and AUCPR scores is similar to the AUCROC scores as shown in Table \ref{cifar10}. It can be noticed that the FPR95 and AUCPR values of the Tiny ImageNet dataset are poor compared to other evaluated datasets that achieve the best possible values. We  relate this behavior to the overlapping object classes and the influence of image resolution as explained in Section \ref{results}.

\begin{table}
  \caption{FPR95 and AUCPR  values of our InFlow model with CIFAR 10 training data as in-distribution samples compared with other OOD detection methods.}
  \label{cifar10sup}
  \centering
  \begin{tabular}{lllllll}
\multicolumn{7}{c}{FPR95}\\
 \toprule   
   Datasets  & InFlow     & Likelihood Ratio     &  LR & ODIN & Outlier exposure & IC  \\
    \midrule
   MNIST                 & 0        &  0.150  &  0       &  0.014     & 0.006     &  0           \\
   FashionMNIST          & 0        &  0.200  &  0       &  0.028     & 0.027     &  0          \\
    SVHN                 & 0        &  0.960  &  0       &   0.155    & 0.076     &  0           \\
    CelebA               & 0        &  0.720  &  0       &  0.181     & 0.517     &  0         \\
     CIFAR 10 (train)    & 0.047    & 0.032   &  0.031   &  0.941     & 0.945     &  0.042       \\
    CIFAR 10 (test)      & 0.049    & 0.035   &  0.035   &  0.950     & 0.950     &  0.050       \\
    Tiny ImageNet        & 0.139    & 0.575   &  0       &  0.412     & 0.076     &  0.075           \\
    Noise                & 0        & 0       &  0       &  0         & 0.008     &  0      \\
    Constant             & 0        & 0       &  0       &  0.413     & 0.001     &  0      \\
    \multicolumn{7}{c}{AUCPR}
       \\ \toprule       
   Datasets  & InFlow     & Likelihood Ratio     &  LR & ODIN & Outlier exposure & IC  \\
    \midrule
   MNIST               & 1           &   0.910  &  0.969    & 0.997       & 0.992   &  0.942         \\
   FashionMNIST        & 1           &  0.909   &  0.914    & 0.994       & 0.976   &  0.849         \\
    SVHN               & 1           &  0.403   & 0.489     & 0.956       & 0.908   &  0.774          \\
    CelebA             & 1           &  0.794   &  0.532    & 0.946       & 0.562   &  0.355       \\
    CIFAR 10 (train)   & 0.567       & 0.511    &  0.503    &  0.509      & 0.170   &  0.503      \\
    CIFAR 10 (test)    & 0.561       & 0.506    & 0.497     & 0.500       & 0.165   &  0.499        \\
    Tiny ImageNet      & 0.515       &  0.129   & 0.498     & 0.950       & 0.935   &  0.126            \\
    Noise              & 1           &   0.189  &   0.888   &  1          & 0.934   &  0.407      \\
    Constant           & 1           &  0.656   &  0.673    &  0.871      & 0.998   &  1     \\
    \bottomrule
  \end{tabular}
\end{table}

\paragraph{FashionMNIST:}We present further results where we evaluate our model trained on the FashionMNIST dataset and inferred on other datasets. Table \ref{fmnist} provides the AUCROC, FPR95, and AUCPR results of our InFlow model when trained with in-distribution FashionMNIST dataset and compared the results with other datasets. Our method can detect the dataset drift and provides robust results for detecting the OOD samples from all evaluated datasets. Additionally, we don't observe the inferior performance of our model while detecting Tiny ImageNet as OOD since the object classes in the Tiny ImageNet dataset are mutually exclusive from the object classes in the FashionMNIST dataset.

\begin{table}
  \caption{AUCROC, FPR95 and AUCPR values of our InFlow model trained on FashionMNIST training data compared with other OOD detection methods. }
  \label{fmnist}
  \centering
  \begin{tabular}{llllll}
  \multicolumn{6}{c}{AUCROC}\\
    \toprule                   
   Datasets  & InFlow     & Likelihood Ratio     &  LR  & Outlier exposure & IC  \\
    \midrule
   MNIST                 & 1       &  0.978            & 1          &    1       &  0.769             \\
  FashionMNIST (train)   & 0.554   &  0.510            & 0.503      &   0.529   &  0.506             \\
   FashionMNIST (test)   & 0.549   &  0.494            & 0.494      &   0.522   &  0.499             \\
    SVHN                 & 1       &  0.981            &  1         &   0.891        &  0.927             \\
    CelebA               & 1       &  0.958            &   1        &   0.823        &  0.378             \\
    CIFAR 10             & 1       &  0.995            &   1        &   0.814        &  0.611             \\
    Tiny ImageNet        & 1       &  1                &   1        &   0.817        &  0.254             \\
    Noise                & 1       &  1                &   1        &   0.890        &  0.531             \\
    Constant             & 1       &  0.935            &   0.716    &   0.999        &   1                \\
    \multicolumn{6}{c}{FPR95}\\
    \toprule                   
    Datasets & InFlow     & Likelihood Ratio     &  LR  & Outlier exposure & IC  \\
    \midrule
   MNIST                   & 0       & 0.175          &   0        &  0    &   0            \\
   FashionMNIST (train)    & 0.040   &  0.971           & 0.019      &   0.012   &  0.048             \\
   FashionMNIST (test)     & 0.044   & 0.975          &  0.025     &   0.015   &   0.050        \\
    SVHN                   & 0       & 0.020          &    0       &  0.575    &   0            \\
    CelebA                 & 0       & 0.053          &    0       &  0.601    &   0.227        \\
    CIFAR 10               & 0       & 0.010          &    0       &  0.637    &   0.075        \\
    Tiny ImageNet          & 0       & 0              &    0       &  0.639    &   0.4          \\
    Noise                  & 0       & 0              &   0        &  0.040    &   0            \\
    Constant               & 0       &  0.075         &  0.25      & 0.001     &   0            \\
    \multicolumn{6}{c}{AUCPR}\\
    \toprule                   
   Datasets  & InFlow     & Likelihood Ratio     &  LR  & Outlier exposure & IC  \\
    \midrule
   MNIST                   & 1       &  0.692         &   1        &   1       &   0.756        \\
   FashionMNIST (train)   & 0.602   &  0.473          & 0.501      &   0.548   &  0.507             \\
   FashionMNIST (test)     & 0.595   &  0.467         & 0.493      &  0.546    &   0.496        \\
    SVHN                   & 1       &  0.487         &  1         & 0.871     &   0.971        \\
    CelebA                 & 1       &  0.525         &  0.999     &  0.834    &   0.612        \\
    CIFAR 10               & 1       &  0.354         &   1        &  0.823    &   0.892        \\
    Tiny ImageNet          & 1       &  0.693         &    1       &  0.780    &   0.408        \\
    Noise                  & 1       &  0.693         &    1       &  0.908    &   0.464        \\
    Constant               & 1       &  0.653         &   0.839    &  0.997    &    1           \\
    \bottomrule
  \end{tabular}
\end{table}

\subsection{Robustness to adversarial attacks}
\label{robustnessadv}

We are interested in examining whether the InFlow model is robust to adversarial attacks. The adversarial attacks are tiny perturbations to the in-distribution samples that are completely hidden from human observation but severely impact the performance of a deep learning model in several real-world applications.
The methods to generate adversarial attacks are commonly targeted to fool supervised learning models. For performing such attacks, we focused on fooling a Reinforcement Learning (RL) agent that is playing Atari games. In Reinforcement learning (RL), multiple actions (labels) might be considered as correct or appropriate. Generated attacks should therefore not only be visible perturbations but ones that worsen the overall performance of the model.

\paragraph{Training RL Agents:} The task of playing Atari games using RL agents is well established in the RL-based literature. Hence, detecting attacks on Atari images acts as a useful evaluation of our model and provides a use case for more complex applications such as autonomous driving. Therefore, in preparation for the adversarial attacks, three agents were trained, each for three different Atari games namely Enduro, RoadRunner, and Breakout. 
They were implemented as Dueling Deep Q-Networks (DQN) \citep{ref42} trained on observations while the optimization was done utilizing the Double DQN \citep{ref41} algorithm.

Throughout a certain amount of time steps, an agent interacts with an environment. In our case, the environment returns a $210 \times 160$ grayscaled image of the game at each step. These so-called observations need to be preprocessed further. To keep recent history, the agent will not only be presented with the current but additional last three images. All images are stacked to create a $210 \times 160 \times 4$ observation on which the agent will base the decision for the next step.

The primary metric to evaluate an agent's performance during training is the average episode reward over 100 episodes. \citep{ref40} provided the scores of a human expert player
for the games. If the average episode reward is significantly higher, the agent was considered
reliable. Our empirical evaluation conveyed that the trained agents performed reliably on their tasks and therefore are suitable attack targets.

\paragraph{Adversarial attack algorithm:}
Now given three such reliable Atari agents, the goal was to find a perturbation vector to be added to the observations, which will be able to fool the RL agent to output erroneous predictions. The perturbations are restricted to stay within a set range of $[-\epsilon,\epsilon]$. The main purpose of the attack was to lower the average episode reward of the agents. It was considered to be successful if: $1)$ the produced adversarial perturbations are unrecognisable for humans; $2)$ they lower the overall performance of the agent, and $3)$ lead to predictions that are unfit for the current state of the environment. We utilized the algorithm of class-discriminative universal adversarial perturbations (CD-UAP) introduced by \citep{ref43} to calculate perturbations that fulfill these criteria. In our case, the perturbations are not universal but input-specific. With this alteration, the algorithm produces a new perturbation for each observation. Although a universal perturbation is more complicated to calculate, it would be easily detected by InFlow if added repeatedly to the in-distribution data.  
\paragraph{Dataset creation:}There were a total of $10,000$ unattacked observations for each of the Atari agents for Breakout, Enduro and RoadRunner. 
We produced adversarially attacked samples with $\epsilon = 0.0008, 0.0009, 0.001, 0.002, 0.003, 0.004$ and $0.005$ which were used as the test samples during inference of the model. It is to be noted that not all of the $10,000$ calculated perturbations were able to pass the above three requirements for our attacks\textemdash  the amount of actual samples for each $\epsilon$ included in the dataset is listed in Table \ref{atarinumimages}. 

\begin{table}[!htb]
  \caption{Number of samples that passed all the criteria for producing a natural adversarial example.}
  \label{atarinumimages}
  \centering
  \begin{tabular}{lllllllll}
    \toprule  
    
      & Unattacked   & 0.0008 &  0.0009 & 0.001 & 0.002  & 0.003 & 0.004 & 0.005 \\
    \midrule
   Breakout          & 10,000  & 3626 & 3735 & 3816 & 4464 & 8216 & 8487 & 10,000  \\
   Enduro            & 10,000  & 6410 & 6730 & 6405 & 8758 & 7105 & 9336 & 10,000  \\
    RoadRunner       & 10,000  & 8700 & 8882 & 9065 & 9867 & 9980 & 9993 & 10,000 \\  
    \bottomrule
  \end{tabular}
\end{table}

\begin{wrapfigure}{r}{0.4\textwidth}
    \centering
    \includegraphics[width=0.4\textwidth]{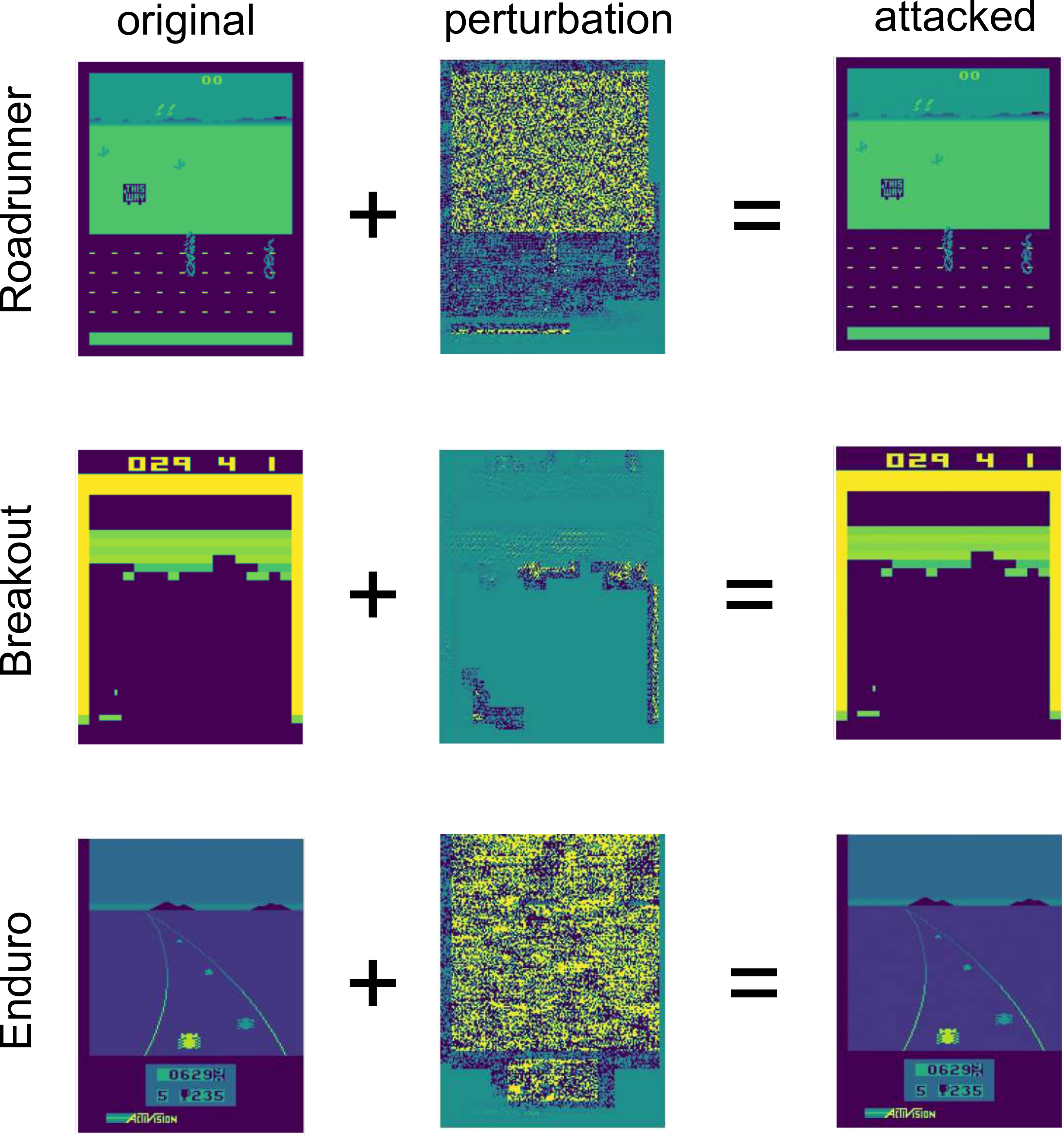}
  \caption{CD-UAP on Breakout, Enduro and RoadRunner.}
  \label{atariimages}
\end{wrapfigure}
\paragraph{Training and Results:}
We trained three separate instances of our InFlow model for each of the games with the original unattacked observations set as the in-distribution samples. Table \ref{ataritable} shows the AUCROC, FPR95 and AUCPR values obtained for the adversarially attacked samples with different values of $\epsilon$. The score reveals that our model is robust in detecting adversarial examples for Breakout and Enduro games where we observe a overall tendency that the scores are better for higher $\epsilon$ as expected. However, for RoadRunner, we do not observe compatible AUCROC scores even for perturbations at $\epsilon = 0.005$. A reason for this behavior could be the higher action space of RoadRunner in comparison to the two other games. Utilizing the CD-UAP algorithm, we determined multiple favorable actions for the current observation and shifted the predictions away from them. For Breakout and Enduro, the different actions are considerably more contradicting and hence the perturbations need to include more distinct features to fool the agent. Additionally, we believe that the created perturbations can be disguised more easily in the detailed images of RoadRunner.

\begin{table}[!htb]
  \caption{AUCROC, FPR95 and AUCPR values of our InFlow model trained on original Atari images as in-distribution and compared with the adversarially attacked images at different $\epsilon$ level of attack with p-value fixed at $\alpha = 0.05$.}
  \label{ataritable}
  \centering
  \begin{tabular}{lllllllll}
     \multicolumn{9}{c}{AUCROC}\\
    \toprule  
    
      & unattacked    & 0.0008      &  0.0009 & 0.001 & 0.002  & 0.003 & 0.004 & 0.005 \\
    \midrule
   Breakout  & 0.524 & 0.985 & 0.994 & 0.983 & 0.991 & 0.977 & 0.998 & 0.973\\
   Enduro  & 0.500 & 0.938 & 0.880 & 0.918 & 0.928 & 0.981 & 0.959 & 0.939 \\
    RoadRunner & 0.524 & 0.598 & 0.629 & 0.617 & 0.604 & 0.606 & 0.639 & 0.654 \\
   \multicolumn{9}{c}{FPR95}\\
    \toprule
         & unattacked     & 0.0008    &  0.0009 & 0.001 & 0.002  & 0.003 & 0.004 & 0.005 \\

    \midrule
   Breakout  & 0.048 & 0.001 & 0 & 0 & 0 & 0 & 0  & 0  \\
   Enduro & 0.050 & 0.007 & 0.011 & 0.009 & 0.006 & 0.001 & 0.003  & 0.003 \\
    RoadRunner & 0.049 & 0.022 & 0.019 & 0.019 & 0.035 & 0.039 & 0.036   & 0.030 \\
       \multicolumn{9}{c}{AUCPR}\\
        \toprule
             & unattacked     & 0.0008      &  0.0009 & 0.001 & 0.002  & 0.003 & 0.004 & 0.005  \\
    \midrule
   Breakout  & 0.549 & 0.971 & 0.980 & 0.971 & 0.981 & 0.982 & 0.992  & 0.983   \\
   Enduro & 0.500 & 0.989 & 0.979 & 0.985 & 0.990 & 0.995 & 0.994  & 0.993 \\
    RoadRunner & 0.556 & 0.761 & 0.788 & 0.781 & 0.782 & 0.776 & 0.802 & 0.810 \\  
        \bottomrule

  \end{tabular}
\end{table}

\subsection{Robustness to visible perturbations}
\label{robustnessnoise}
\begin{figure}
 \centering
  \includegraphics[width=0.98\linewidth]{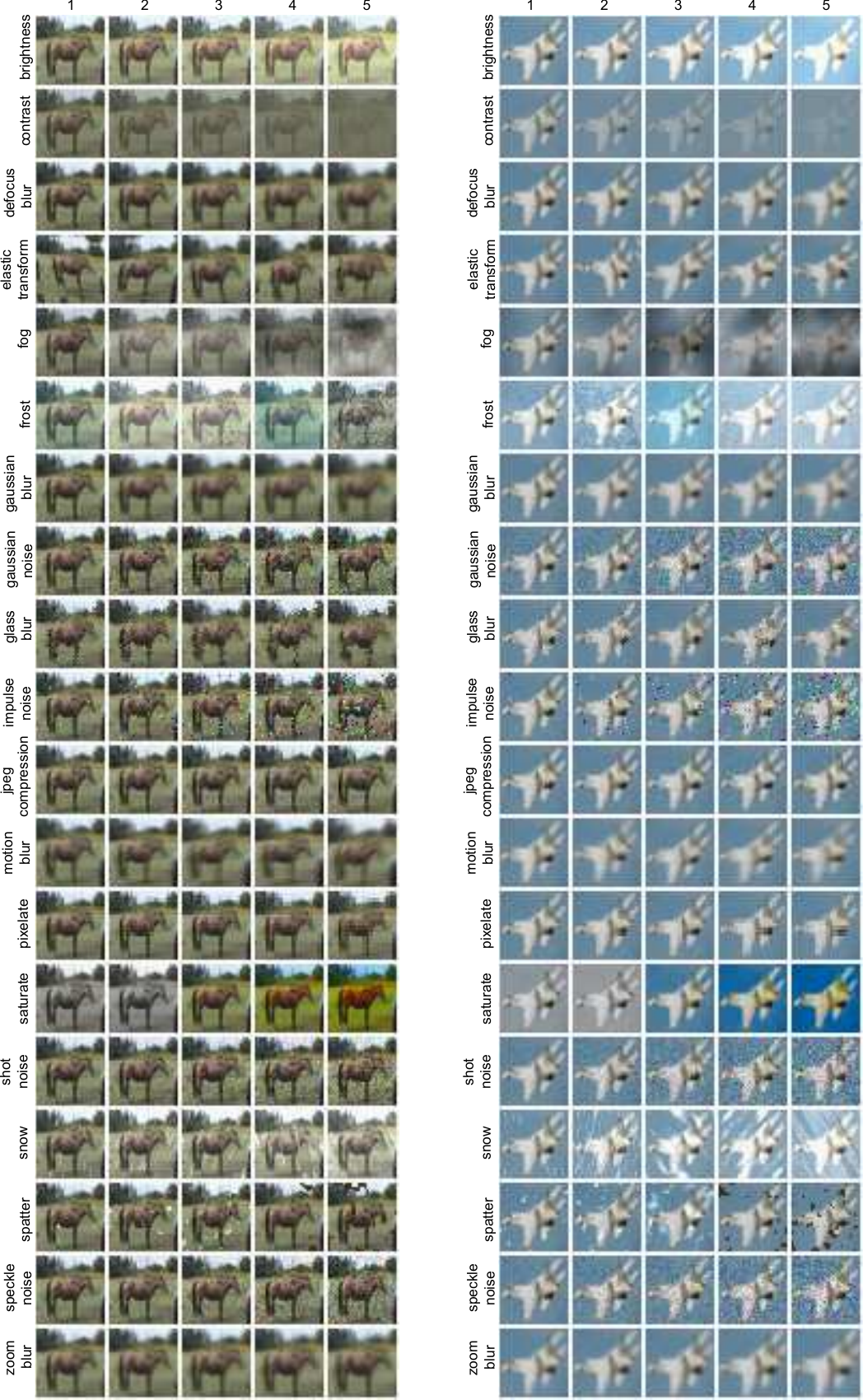}
  \caption{The different perturbations on two CIFAR 10 images with increasing level of severity.}
  \label{perturbationimages}
\end{figure}

\paragraph{Data generation:} To evaluate the robustness of our InFlow model on visibly perturbed images, we generated corrupted versions of CIFAR 10 test samples on 19 different types of perturbation at five separate severity levels as proposed in \citep{ref44}. The 19 different perturbation types were chosen since several existing ML models showcase instability for accurately predicting the object classes after the samples undergo such perturbations.  Figure \ref{perturbationimages} shows two CIFAR 10 test examples with five different levels of perturbation severity applied to them. These perturbation types can be categorized into four broad categories\textemdash Noise, Blur, Weather and Digital effects\textemdash that significantly cover a broad spectrum of real-world perturbations. 

\begin{table}[!htb]
  \caption{AUCROC values of our InFlow model trained on  CIFAR 10 training images as in-distribution and evaluated with different types of visible perturbations at increasing severity levels.}
  \centering
  \renewcommand{\arraystretch}{1.2}
  \begin{tabular}{p{2.7cm}ccccc}
    \toprule
    Perturbation Type & Severity 1 & Severity 2 &
    Severity 3 & Severity 4 & Severity 5\\
    \midrule
    Gaussian Noise    & 0.229 & 0.260 & 0.705 & 0.851 & 1.0  \\
    Impulse Noise     & 0.314 & 0.566 & 0.901 & 1.0   & 1.0  \\ 
    Shot Noise        & 0.343 & 0.272 & 0.358 & 0.571 & 0.951  \\ 
    Speckle Noise     & 0.344 & 0.239 & 0.317 & 0.584 & 1.0  \\
    Defocus Blur      & 0.541 & 0.628 & 0.909 & 0.978 & 1.0  \\ 
    Gaussian Blur     & 0.540 & 0.909 & 0.978 & 1.0 & 1.0  \\
    Glass Blur        & 0.514 & 0.466 & 0.565 & 0.464 & 0.601  \\ 
    Motion Blur       & 0.602 & 0.955 & 1.0   & 1.0   & 1.0  \\ 
    Zoom Blur         & 0.815 & 0.910 & 0.977 & 0.978 & 0.978  \\ 
    Snow              & 1.0 & 1.0 & 1.0 & 1.0 & 1.0  \\ 
    Spatter           & 0.560 & 0.809 & 1.0 & 0.593 & 0.806  \\
    Frost             & 1.0 & 1.0 & 1.0 & 1.0 & 1.0  \\ 
    Fog               & 1.0 & 1.0 & 1.0 & 1.0 & 1.0  \\ 
    Brightness        & 1.0 & 1.0 & 1.0 & 1.0 & 1.0  \\  
    Contrast          & 1.0 & 1.0 & 1.0 & 1.0 & 1.0  \\
    Saturate          & 1.0 & 1.0 & 1.0 & 1.0 & 1.0  \\
    Elastic Transform & 0.571 & 0.649 & 0.791 & 0.884 & 0.860  \\ 
    Pixelate          & 0.500 & 0.525 & 0.550 & 0.547 & 0.729  \\ 
    JPEG Compression  & 0.485 & 0.509 & 0.509 & 0.509 & 0.491  \\ 
    \bottomrule
  \end{tabular}
  \label{perturbationscore}
\end{table}

\paragraph{Training:} We trained our InFlow model on original CIFAR 10 training images and tested the efficiency of the model for detecting the increasing severity levels of corruptness. 
For training our model, we fixed the significance p-value of our attention mechanism at $\alpha = 0.05$ while keeping the hyperparameters as set in Appendix \ref{trainingdetails} and used the encoder architecture as shown in Table \ref{encoder}. We then calculated the AUCROC scores as shown in Table \ref{perturbationscore}, which were obtained for all 19 perturbation types with increasing severity levels while keeping CIFAR 10 training samples as in-distribution. 

\paragraph{Results:} Observably, our model can detect OOD samples related to weather effects such as frost, fog, and brightness at all severity levels with perfect AUCROC scores. For the spatter effect, we notice a different color pattern at severity levels 4 and 5, in contrast to the pattern at severity levels 1 to 3. This behavior explains the drop in the performance of the InFlow model with an increase in severity from level 3 to level 4. For perturbations in the noise and blur categories, an increasing level of corruptness results in increased  AUCROC scores as the more severely perturbed test samples get significantly further away from being in-distribution. However, for shot noise, speckle noise, and glass blur, there is a dip in AUCROC scores when the severity is increased from 1 to 2. We argue that at severity level 2, the perturbation resembled in-distribution samples. Overall, it can be presumed that our model is robust in detecting several types of visible perturbations on in-distribution samples as OOD and its performance improves as the severity of the corruption is increased. 

\subsection{Visualization of sub-network activations and latent space}
\label{visualization}

\begin{figure}[!h]
 \centering
  \includegraphics[width=\linewidth]{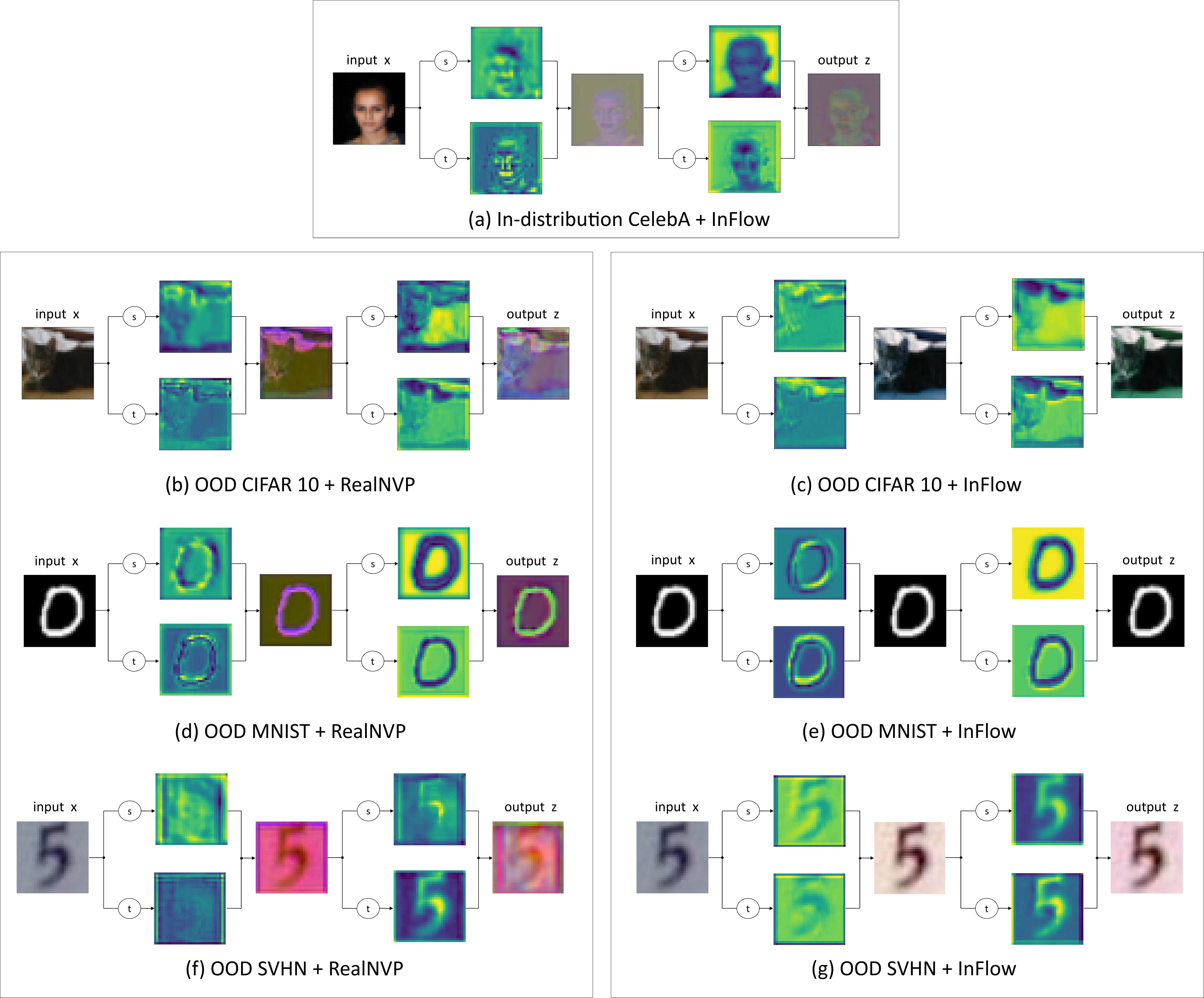}
  \caption{The figure visualizes the input $x$, the activations of the sub-networks $s$ and $t$, the activations of the coupling blocks and the output $z$ in latent space of in-distribution and several OOD datasets obtained from the RealNVP model and our InFlow model at K = 2. }
  \label{visimages}
\end{figure}

We visually interpreted the variations in the behavior of in-distribution and OOD samples when using our model compared to RealNVP based flow model \citep{ref2}. We performed several visualizations of sub-network activations, output at each coupling block, and the latent space. We used a total of two coupling blocks ($K=2$) for both RealNVP and our model and adopted non-shared weights for $s$ and $t$ activations at each coupling block. Figure \ref{visimages} shows the visualization results for RealNVP as well as our InFlow model when compared with in-distribution CelebA and other OOD datasets. 

It can be observed from Figure \ref{visimages} (a) that the InFlow model transforms the in-distribution CelebA samples $x$ into a complex latent space $z$ after training. Additionally, Figures \ref{visimages} (b), (d), and (f) reveal that the RealNVP model with no attention mechanism in place also transforms the OOD datasets into a much more complex distribution in the latent space. This presents the visual proof that the RealNVP model learns the local pixel interactions of the input space for both in-distribution and OOD samples due to which it cannot distinguish between the semantic information from an in-distribution sample and an OOD outlier.  As a consequence, the RealNVP model increases the log-likelihood of both in-distribution and OOD samples. Figures \ref{visimages} (c), (e), and (g) show the visualization of our InFlow model for different OOD datasets. It is noticeable that our model reproduces the semantic input features into its latent space and a color change occurs for the output of the coupling blocks as a consequence of the permutation. For the grayscale images like MNIST, the color change is not visible since all the three channels are the same grayscale image. 
Therefore, the depiction in Figure \ref{visimages} provides visual evidence that the InFlow model directly transfers the input in 
its latent space for OOD samples while it was able to transform the in-distribution samples into a much more complex distribution.

\subsection{InFlow architecture vs Robustness}
\label{modelarchitecturewithrobustness}
To analyze the effect of altering the architecture of our model in terms of OOD detection performance, we defined two strategies. The first strategy deals with evaluating the effect of increasing the number of coupling blocks in the model on the OOD detection performance. The second strategy focuses on estimating the effect of jointly or separately learning the weights of $s$ and $t$ sub-networks. Therefore, we trained four separate instances of the InFlow model with the CelebA training images as the in-distribution samples. In both the joint and non-joint learning settings, we evaluated two and four coupling blocks with $K=2$ and $K=4$ respectively.

\begin{figure}[!htb]
\begin{subfigure}{.32\textwidth}
  \includegraphics[width=\linewidth]{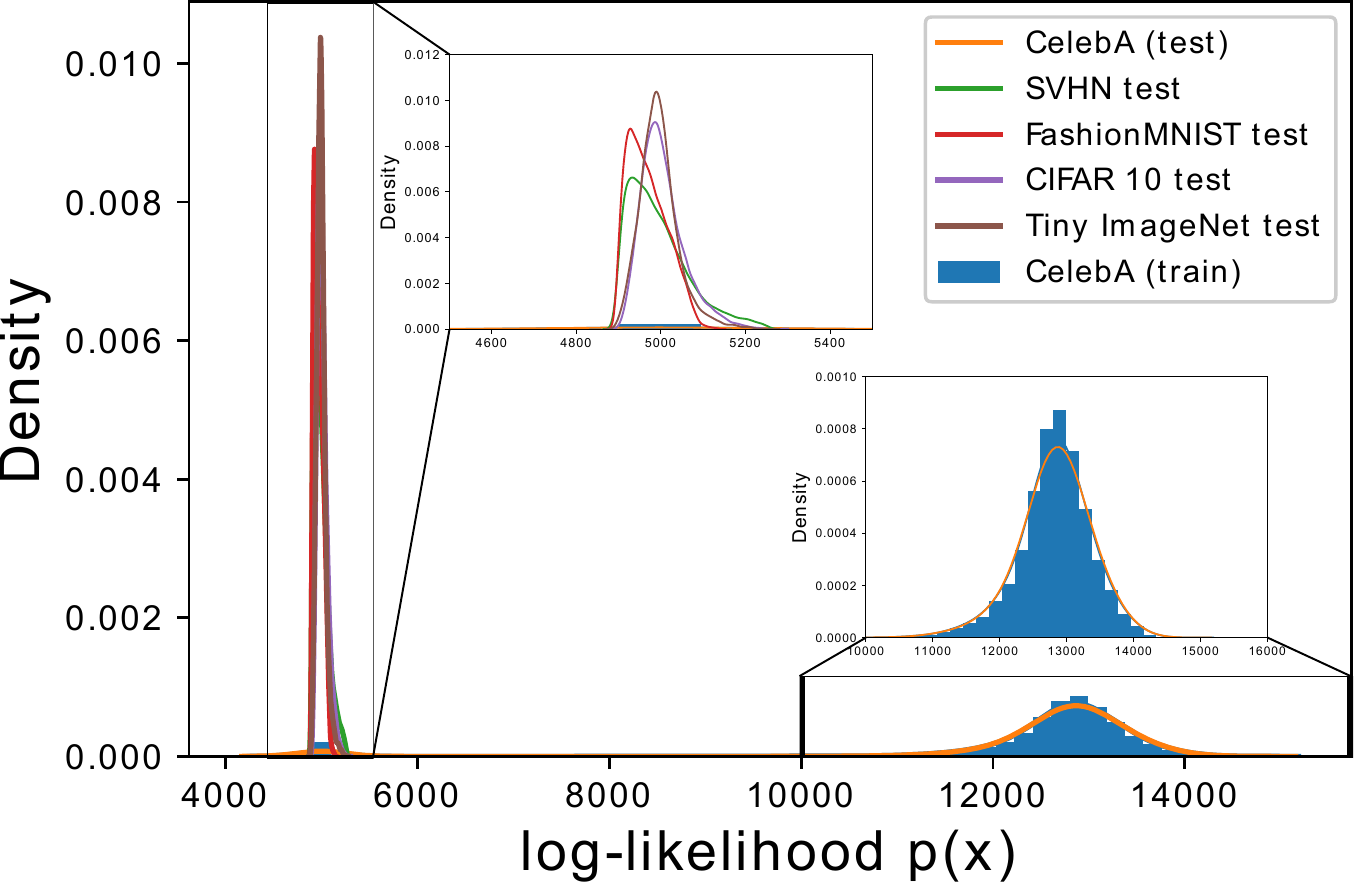}
  \caption{non-shared (K = 4) }
\end{subfigure}%
\begin{subfigure}{.32\textwidth}
  \includegraphics[width=\linewidth]{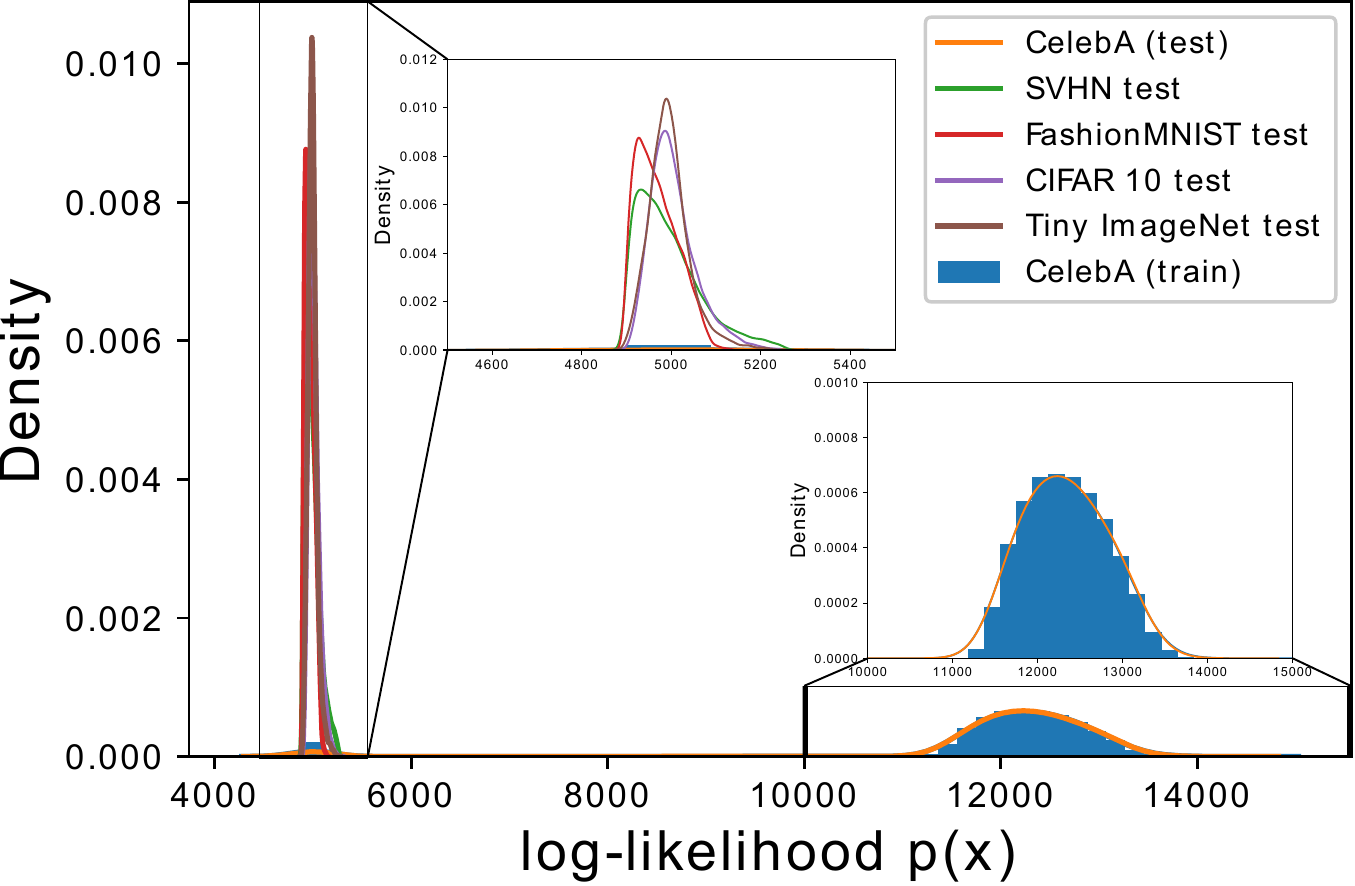}
  \caption{shared (K = 2)}
\end{subfigure}%
\begin{subfigure}{.32\textwidth}
  \includegraphics[width=\linewidth]{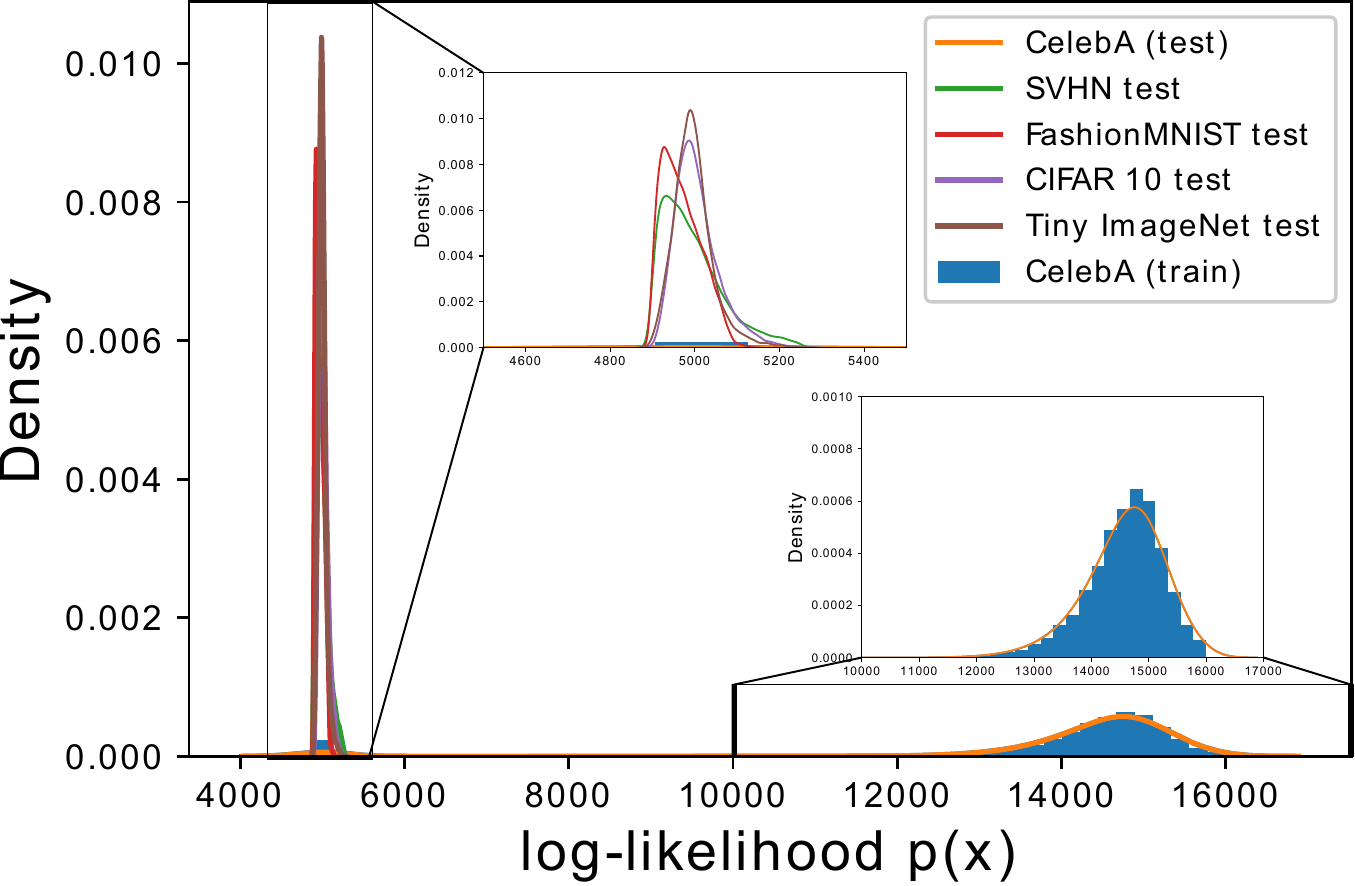}
  \caption{shared (K = 4)}
\end{subfigure}%
\\

\caption{(a) shows the histogram of log-likelihoods with non-shared weights of $s$ and $t$ sub-networks at K = 4. (b) shows the histogram of log-likelihoods with shared weights of $s$ and $t$ sub-networks at K = 2. (c) shows the histogram of log-likelihoods with shared weights of $s$ and $t$ sub-networks at K = 4. }
\label{archi}
\end{figure}

\begin{table}[!htb]
  \caption{AUCROC values of our InFlow model trained on CelebA images as in-distribution and compared with several OOD datasets at different number of coupling blocks.}
  \label{differentlayers}
  \centering
  \begin{tabular}{lllll}
    \toprule  
      & K = 4 (non-shared) & K = 2 (shared) & K = 4 (shared) \\
    \midrule
  MNIST                 &   1           &     1     &       1          \\
  FashionMNIST          &   1           &    1      &        1          \\  
  SVHN                  &   1           &   1       &        1          \\
    CelebA (train)         &  0.522        &  0.527        &     0.526              \\ 
  CelebA (test)         &  0.523        &  0.520        &     0.523              \\ 
   CIFAR10              &  1        &   1       &    1               \\
     Tiny ImageNet      &  1        &    1      &    1                \\
       Noise            &   1           &    1      &         1           \\
       Constant         &  1            &    1      &         1           \\
    \bottomrule 
  \end{tabular}
\end{table}

 Figure \ref{archi} shows the histogram of log-likelihoods of the InFlow obtained after training it in the four mentioned architectural settings, and Table \ref{differentlayers} shows the AUCROC values of these settings at $\alpha = 0.05$. The AUCROC values for the non-shared weights of $s$ and $t$ sub-networks with $K = 2$ coupling blocks and p-value $\alpha = 0.05$  can be found in Table \ref{alpha}. The results reveal that in each of the modified architectural settings, our InFlow model was able to assign lower log-likelihood to OOD datasets compared to the in-distribution CelebA dataset. Therefore, we can conclude that the internal architecture of the InFlow model does not affect the performance w.r.t OOD detection. Hence, this provides a significant empirical observation that should prevent further study on improving the design of normalizing flows, particularly for OOD detection.



\begin{thebibliography}{999}

\bibitem[Dinh et al.(2015)]{ref1}
 Laurent Dinh, David Krueger, and Yoshua Bengio. NICE: Non-linear Independent Components Estimation. In {\it International Conference on Learning Representations (ICLR)}, 2015.

\bibitem[Dinh et al.(2017)]{ref2}
Laurent Dinh, Jascha S. Dickstein, and Samy Bengio. Density estimation using Real NVP. In {\it International Conference on Learning Representations (ICLR)}, 2017.

\bibitem[Sorrenson et al.(2020)]{ref3}
Peter Sorrenson, Carsten Rother, and Ullrich Köthe. Disentanglement by nonlinear ICA with general incompressible-flow networks (GIN). In {\it International Conference on Learning Representations (ICLR)}, 2020.

\bibitem[Kingma et al.(2018)]{ref4}
 Durk P. Kingma, and Prafulla Dhariwal. Glow: Generative Flow with Invertible 1x1 Convolutions. In {\it Advances in Neural Information Processing Systems (NIPS)}, 2018.

\bibitem[Grathwohl et al.(2019)]{ref5}
 Will Grathwohl, Ricky T. Q. Chen, Jesse Bettencourt, Ilya Sutskever, and David Duvenaud. FFJORD: Free-form Continuous Dynamics for Scalable Reversible Generative Models. In {\it International Conference on Learning Representations (ICLR)}, 2019.


\bibitem[Durkan et al.(2019)]{ref6}
 Conor Durkan, Artur Bekasov, Iain Murray, and George Papamakarios. Neural Spline Flows. In {\it Advances in Neural Information Processing Systems (NIPS)}, 2019.
 
\bibitem[Nalisnick et al.(2019)]{ref7}
Eric Nalisnick, Akihiro Matsukawa, Yee W. Teh, Dilan Gorur, and Balaji Lakshminarayanan. Do deep generative models know what they don’t know?. In {\it International Conference on Learning Representations (ICLR)}, 2019.

  
\bibitem[Kirichenko et al.(2020)]{ref8}
Polina Kirichenko, Pavel Izmailov and Andrew G. Wilson. Why Normalizing Flows Fail to Detect Out-of-Distribution Data. In {\it Advances in Neural Information Processing Systems (NIPS)}, 2020.

\bibitem[Gretton et al.(2012)]{ref9}
Arthur Gretton, Karsten M. Borgwardt, Malte J. Rasch, Bernhard Schölkopf, and Alexander Smola. A Kernel Two-Sample Test. {\it Journal of Machine Learning Research (JMLR)}, 2012.


\bibitem[Nguyen et al.(2012)]{ref10}
 Anh Nguyen, Jason Yosinski, and Jeff Clune. Deep Neural Networks are Easily Fooled: High Confidence Predictions for Unrecognizable Images. In {\it IEEE Conference on Computer Vision and Pattern Recognition (CVPR)}, 2015.


\bibitem[Ren et al.(2019)]{ref11}
Jie Ren, Peter J. Liu, Emily Fertig, Jasper Snoek, Ryan Poplin, Mark Depristo, Joshua Dillon, and Balaji Lakshminarayanan. Likelihood ratios for out-of-distribution detection. In {\it Advances in Neural Information Processing Systems (NIPS)}, 2019.


\bibitem[Liang et al.(2018)]{ref12}
 Shiyu Liang, Yixuan Li, and R. Srikant. Enhancing The Reliability of Out-of-distribution Image Detection in Neural Networks.  In {\it International Conference on Learning Representations (ICLR)}, 2018.

\bibitem[DeVries et al.(2018)]{ref13} 
Terrance DeVries, and Graham Wr. Taylor. Learning Confidence for Out-of-Distribution Detection in Neural Networks. {\it arXiv preprint arXiv:1802.04865}, 2018.

\bibitem[Hendrycks et al.(2018)]{ref14} 
Dan Hendrycks, Mantas Mazeika, and Thomas Dietterich. Deep Anomaly Detection with Outlier Exposure. In {\it International Conference on Learning Representations (ICLR)}, 2019.

\bibitem[Hendrycks et al.(2018)]{ref15} 
Dan Hendrycks, and Kevin Gimpel. A Baseline for Detecting Misclassified and Out-of-Distribution Examples in Neural Networks. In {\it International Conference on Learning Representations (ICLR)}, 2017.

\bibitem[Lee et al.(2018)]{ref16} 
Kimin Lee, Kibok Lee, Honglak Lee, and Jinwoo Shin. A Simple Unified Framework for Detecting Out-of-Distribution Samples and Adversarial Attacks. In {\it Advances in Neural Information Processing Systems (NIPS)}, 2018.

\bibitem[Chen et al.(2020)]{ref17} 
 Jiefeng Chen, Yixuan Li, Xi Wu, Yingyu Liang, and Somesh Jha. Robust Out-of-distribution Detection for Neural Networks. {\it arXiv preprint arXiv:2003.09711}, 2020. 

\bibitem[Ardizzone et al.(2020)]{ref18} 
 Lynton Ardizzone, Radek Mackowiak, Carsten Rother, and Ullrich Köthe. Training Normalizing Flows with the Information Bottleneck for Competitive Generative Classification. In {\it Advances in Neural Information Processing Systems (NIPS)}, 2020.

\bibitem[Ardizzone et al.(2019)]{ref19} 
 Lynton Ardizzone, Jakob Kruse, Sebastian Wirkert, Daniel Rahner, Eric W. Pellegrini, Ralf S. Klessen, Lena Maier-Hein, Carsten Rother, and Ullrich Köthe. Analyzing inverse problems with invertible neural networks. In {\it International Conference on Learning Representations (ICLR)}, 2019.

\bibitem[Lee et al.(2018)]{ref20} 
 Kimin Lee, Honglak Lee, Kibok Lee, and Jinwoo Shin. Training Confidence-calibrated Classifiers for Detecting Out-of-Distribution Samples. In {\it International Conference on Learning Representations (ICLR)}, 2018. 

\bibitem[Hendrycks et al.(2019)]{ref21} 
 Dan Hendrycks, Mantas Mazeika, Saurav Kadavath, and Dawn Song. Using Self-Supervised Learning Can Improve Model Robustness and Uncertainty. In {\it Advances in Neural Information Processing Systems (NIPS)}, 2019.

\bibitem[Serrà et al.(2019)]{ref22} 
Joan Serrà, David Álvarez, Vicenç Gómez, Olga Slizovskaia, José F. Núñez, and Jordi Luque. Input complexity and out-of-distribution detection with likelihood-based generative models. In {\it International Conference on Learning Representations (ICLR)}, 2019.

\bibitem[Kobyzev et al.(2020)]{ref23} 
 Ivan Kobyzev, Simon Prince, and Marcus Brubaker. Normalizing Flows: An Introduction and Review of Current Methods. {\it IEEE Transactions on Pattern Analysis and Machine Intelligence}, 2020.


\bibitem[Xiao et al.(2020)]{ref25} 
 Zhisheng Xiao, Qing Yan, and Yali Amit. Likelihood Regret: An Out-of-Distribution Detection Score For Variational Auto-encoder. In {\it Advances in Neural Information Processing Systems (NIPS)}, 2020.

\bibitem[Morningstar et al.(2021)]{ref26} 
Warren R. Morningstar, Cusuh Ham, Andrew G. Gallagher, Balaji Lakshminarayanan, Alexander A. Alemi, and Joshua V. Dillon. Density of States Estimation for Out-of-Distribution Detection. In {\it International Conference on Artificial Intelligence and Statistics (AISTATS)}, 2021.

\bibitem[Chen et al.(2021)]{ref27} 
 Jiefeng Chen, Yixuan Li, Xi Wu, Yingyu Liang, and Somesh Jha. Informative Outlier Matters: Robustifying Out-of-distribution Detection Using Outlier Mining. In {\it International Conference on Learning Representations (ICLR)}, 2021.

\bibitem[Zisselman et al.(2020)]{ref28} 
 Ev Zisselman, and Aviv Tamar.
Deep Residual Flow for Out of Distribution Detection. In {\it The IEEE Conference on Computer Vision and Pattern Recognition (CVPR)}, 2020. 

\bibitem[Rabanser et al.(2019)]{ref29} 
 Stephan Rabanser, Stephan Günnemann, and Zachary C. Lipton. Failing Loudly: An Empirical Study of Methods for Detecting Dataset Shift. In {\it Advances in Neural Information Processing Systems (NIPS)}, 2019.

\bibitem[Nalisnick et al.(2020)]{ref30} 
 Eric Nalisnick, Akihiro Matsukawa, Yee Whye Teh, and Balaji Lakshminarayanan. Detecting Out-of-Distribution Inputs to Deep Generative Models Using Typicality. In {\it International Conference on Learning Representations (ICLR)}, 2020.


\bibitem[Guo et al.(2018)]{ref32} 
 Chuan Guo, Mayank Rana, Moustapha Cisse, and Laurens van der Maaten. Countering adversarial images using input Transformations. In {\it International Conference on Learning Representations (ICLR)}, 2018.

\bibitem[Choi et al.(2018)]{ref33} 
 Hyunsun Choi, Eric Jang, Alexander A. Alemi. WAIC, but Why? Generative Ensembles for Robust Anomaly Detection. {\it arXiv preprint arXiv:1810.01392}, 2019. 
 
 \bibitem[Liu et al.(2018)]{ref34} 
 Ziwei Liu, Ping Luo, Xiaogang Wang, and Xiaoou Tang.  Deep learning face attributes in the wild.  In {\it Proceedings of the IEEE international conference on computer vision (ICCV)}, 2015.
 
\bibitem[LeCun et al.(2010)]{ref35} 
Yann LeCun, Corinna Cortes, and Christopher J. Burges. Mnist handwritten digit database. 2010.

\bibitem[Xiao et al.(2017)]{ref36}
Han Xiao,  Kashif Rasul, and Roland Vollgraf. Fashion-mnist:  a novel image dataset for benchmarking machine learning algorithms. {\it arXiv preprint arXiv:1708.07747}, 2017.

\bibitem[Netzer et al.(2011)]{ref37}
Yuval Netzer, Tao Wang, Adam Coates, Alessandro Bissacco, Bo Wu, and Andrew Y. Ng. Reading digits in natural images with unsupervised feature learning. In {\it  NIPS Workshop on Deep Learning and Unsupervised Feature Learning}, 2011.

\bibitem[Krizhevsky et al.(2019)]{ref38}
Alex Krizhevsky, Geoffrey Hinton. Learning multiple layers of features from tiny images. 2009

\bibitem[Pouransari et al.(2019)]{ref39}
 Tiny ImageNet Visual Recognition Challenge. \url{https://tiny-imagenet.herokuapp.com/}.


\bibitem[Mnih et. al.(2015)]{ref40}
Volodymyr Mnih, Koray Kavukcuoglu, David Silver, Andrei A. Rusu, Joel Veness, Marc G. Bellemare et. al. Human-level control through deep reinforcement learning. 
 {Nature}, 2015.
 
\bibitem[Hasselt et. al.(2016)]{ref41} 
Hado van Hasselt, Arthur Guez, and David Silver. Deep reinforcement learning with double
Q-Learning. In {\it AAAI Conference on Artificial Intelligence (AAAI)}, 2016.

\bibitem[Wang et. al.(2016)]{ref42}
Ziyu Wang, Tom Schaul, Matteo Hessel, Hado van Hasselt, Marc Lanctot, and Nando de Freitas. Dueling
Network Architectures for Deep Reinforcement Learning. In {\it 33rd International
Conference on Machine Learning  (ICML)}, 2016. 

\bibitem[Zhang et. al.(2020)]{ref43}
Chaoning Zhang, Philipp Benz, Tooba Imtiaz, and In So Kweon. CD-UAP: Class Discriminative Universal Adversarial Perturbation. In {\it AAAI Conference on Artificial Intelligence (AAAI)}, 2020.

\bibitem[Hendrycks et. al.(2019)]{ref44}
Dan Hendrycks, Thomas Dietterich. Benchmarking Neural Network Robustness to Common Corruptions and Perturbations. 
In {\it International Conference on Learning Representations (ICLR)}, 2019.

\bibitem[Lakshminarayanan et. al.(2017)]{ref45}
Balaji Lakshminarayanan, Alexander  Pritzel, and Charles Blundell.   Simple and scalable predictive uncertainty estimation using deep ensembles.  In {\it Advances in Neural Information Processing Systems (NIPS)}, 2017. 

\bibitem[DeVries et. al.(2018)]{ref46}
Terrance DeVries and Graham W Taylor. Learning confidence for out-of-distribution detection in neural networks. {\it arXiv preprint arXiv:1802.04865}, 2018.

\bibitem[Akcay et. al.(2018)]{ref47}
Samet Akcay, Amir A. Abarghouei, and Toby P. Breckon. Ganomaly: Semi-supervised anomaly detection via adversarial training.  In {\it Asian Conference on Computer Vision (ACCV)}, 2018.

\end{thebibliography}
\end{document}